\documentclass[sn-mathphys]{sn-jnl}

\usepackage{gensymb} 
\usepackage{textcomp} 
\usepackage{pifont} 
\usepackage{numprint}
\usepackage{adjustbox}
\usepackage{amsmath,amsthm,amssymb,amsfonts, scrextend}
\usepackage{algorithm}
\usepackage{tabularx} 
\usepackage{array} 

\usepackage{hyperref}

\begin{document}

\title[Critical Assesment]{
A Critical Assessment of Modern Generative Models' Ability to Replicate Artistic Styles}

\newcommand{\E}{\mathbb{E}}

\author*[1]{\fnm{Andrea} \sur{Asperti}}\email{andrea.asperti@unibo.it}

\author[2]{\fnm{Franky} \sur{George}}\email{f.george-2021@hull.ac.uk}

\author[1]{\fnm{Tiberio} \sur{Marras}}\email{tiberio.marras@studio.unibo.it}

\author[1]{\fnm{Razvan Ciprian} \sur{Stricescu}}\email{razvancipr.stricescu@studio.unibo.it}

\author[1]{\fnm{Fabio} \sur{Zanotti}}\email{fabio.zanotti2@studio.unibo.it}

\affil*[1]{\orgdiv{Department of Informatics: Science and Engineering (DISI)}, \orgname{University of Bologna}, \orgaddress{\street{Mura Anteo Zamboni 7}, \city{Bologna}, \postcode{40126},
\country{Italy}
}}

\affil*[2]{\orgdiv{DAIM, Computer Science}, \orgname{University of Hull}, \orgaddress{\street{Cottingham Rd}, \city{Hull}, \postcode{HU6 7RX},
\country{United Kingdom}
}}

\abstract{
In recent years, advancements in generative artificial intelligence have led to the development of sophisticated tools capable of mimicking diverse artistic styles, opening new possibilities for digital creativity and artistic expression. This paper presents a critical assessment of the style replication capabilities of contemporary generative models, evaluating their strengths and limitations across multiple dimensions. We examine how effectively these models reproduce traditional artistic styles while maintaining structural integrity and compositional balance in the generated images.

The analysis is based on a new large dataset of 
AI-generated works imitating artistic styles of the past,
holding potential for a wide range of applications:
the ``AI-pastiche" dataset.

The study is supported by extensive user surveys, collecting diverse opinions on the dataset 
and investigation both technical and aesthetic challenges, including the ability to generate outputs that are realistic and visually convincing, the versatility of models in handling a wide range of artistic styles, and the extent to which they adhere to the content and stylistic specifications outlined in prompts.

This paper aims to provide a comprehensive overview of the current state of generative tools in style replication, offering insights into their technical and artistic limitations, potential advancements in model design and training methodologies, and emerging opportunities for enhancing digital artistry, human-AI collaboration, and the broader creative landscape.}


\maketitle

\section{Introduction}
Generative AI has rapidly expanded into creative fields, transforming how visual art is produced, modified, and experienced. Early breakthroughs, such as StyleGAN \cite{StyleGAN2,StyleGAN-T}, laid the foundation for high-quality image synthesis, but the field has since been revolutionized by the rapid rise of diffusion-based models \cite{DDPM,DDIM,stable-diffusion}. These newer techniques have significantly enhanced the ability to generate realistic images, mimic artistic styles \cite{SGMD,ArtBank}, and even create entirely new visual compositions \cite{Dalle2,Imagen}, establishing diffusion models as the dominant paradigm in generative artistry. Among these capabilities, style replication has emerged as a key area of interest, allowing users to apply diverse historical and modern artistic styles to AI-generated images \cite{FeaST,Style_injection,Step-aware}. This technology enables greater artistic expression and personalization, bridging the gap between computational creativity and traditional artistry and empowering artists, designers, and hobbyists to explore and reinterpret visual styles in ways that were previously highly specialized or time-intensive \cite{designing-barros,GameDesign,inspiring-Haase,craft-edu}.


The purpose of this study is to provide a critical assessment of the capabilities and limitations of current generative tools in effectively replicating styles. By examining both the technical performance and aesthetic outcomes of these tools, the study aims to highlight their strengths, identify areas where they fall short, and offer insights into the potential improvements needed to enhance their application in creative fields. 

Specifically, we compared twelve modern generative models comprising \href{DallE 3}{https://openai.com/index/dall-e-3/}, \href{Stable Diffusion 1.5}{https://huggingface.co/stable-diffusion-v1-5/stable-diffusion-v1-5}, \href{Stable Diffusion 3.5 large}{https://stabledifffusion.com/tools/sd-3-5-large}, \href{Flux 1.1 Pro}{https://flux1.ai/flux1-1}, \href{Flux 1 Schnell}{https://fluxaiimagegenerator.com/flux-schnell}, \href{Omnigen}{https://omnigenai.org/}, \href{Ideogram}{https://ideogram.ai/login}, \href{Kolors 1.5}{https://klingai.com/text-to-image/new}, \href{Firefly Image 3}{https://firefly.adobe.com/}, \href{Leonardo Pheonix}{https://leonardo.ai}, \href{Midjourney V6.1}{https://www.midjourney.com/imagine} and \href{Auto-Aesthetics v1}{https://neural.love/blog/auto-aesthetics-v1-ai-art-revolution}. 

The models were compared using 73 uniform prompts that span a broad range of painting styles from the past five centuries. This resulted in the creation of a
large supervised dataset of AI-generated artworks: the
AI-pastiche dataset. This dataset not only provides labeled examples of AI-generated images but also offers a valuable resource for advancing research in areas such as deepfake detection, digital forensics, and the ethical study of AI-generated content. By supplying a controlled, high-quality set of deepfakes, the dataset aids in training and testing models for improved detection accuracy, robustness against manipulation, and broader exploration of generative AI capabilities across fields ranging from security to digital art.

The quality of the generated images was evaluated based on two criteria: the ability of the models to faithfully replicate human-crafted artwork and their capacity to adhere to the style and content specified in the prompts. The first criterion was assessed through a public survey in which participants were asked to distinguish between human-created and AI-generated images. The second criterion, which involved a per-prompt comparison of the samples generated by different models, was evaluated directly by members of our team along with a few additional volunteers.

Our results reveal that while modern generative models demonstrate remarkable artistic capabilities, they still encounter significant challenges in faithfully replicating historical styles. Rather than a lack of detail, hyperrealism emerges as the primary obstacle—AI-generated images often display excessive sharpness and unnatural precision, making them visually striking but historically inconsistent. According to our evaluation, state-of-the-art models successfully produce images that non-expert users misidentify as human-created in less than 30\% of cases, highlighting the persistent gap between AI-generated and traditionally crafted artworks.

This work is part of a larger and ambitious project that aims to assess whether Large Language Models possess an aesthetic sense and, if so, to identify the aesthetic principles that guide their preferences. This investigation represents a significant advancement in understanding the emergent abilities of LLMs \cite{emergent,emergent_mirage,loss_perspective,Angelo_TOM} and their social implications.
In evaluating the aesthetic sense of LLMs, it is essential to bypass any potential familiarity the models may have with specific artworks, as this could allow them to draw on pre-existing evaluations or learned data. By using a dataset of fictional or AI-generated artworks, such as the one created as part of this study, we can ensure that LLMs rely solely on the information provided in the dataset, thus offering a more controlled evaluation of their aesthetic judgment.

Summing up, our work makes two major contributions:
\begin{itemize}
\item we created a large, richly annotated dataset of AI-generated images, suitable for a wide range of applications;
\item we conducted an in-depth evaluation of some of the most widely used generative models, assessing their style-transfer capabilities through suitable surveys.
\end{itemize}

The article has the following structure. In Section~\ref{sec:methodology}, we describe our methodology, the way the dataset has been created, the selection of models and the way surveys have been
formulated and conducted. Section~\ref{sec:dataset} gives
a detailed description of the dataset, and the relative metadata.
In Section~\ref{sec:surveys} we give a detailed description 
about the surveys, the target audience, and the frameworks used to publish and collect data.
Section~\ref{sec:results} describes the results of the
evaluation. An in-depth discussion of some of the main critical aspects of the style-transfer capabilities of generative tools is given in Section~\ref{sec:discussion}. 
In Section~\ref{sec:conclusions}, we offer a few ideas for
future developments, and outlies some possible applications 
of our dataset.

\section{Related works}

AI-driven artistic style transfer has grown significantly in recent years, driven by advances in deep learning and generative models. Several works have explored the capabilities, limitations, and applications of AI-generated imagery. Our work contributes with a comprehensive evaluation of multiple generative models, emphasizing their adherence to artistic style and prompt fidelity.

Early works such as Gatys et al. \cite{Gatys_neural_representation} laid the foundation for neural style transfer, introducing methods that blend content and style representations from convolutional neural networks. Subsequent research expanded on these concepts, improving efficiency and control over style application \cite{Huang_style_transfer_normalization}. More recently diffusion-based models have demonstrated superior results in high-fidelity artistic synthesis, allowing for more nuanced style adaptation. Our study build upon these advancements but diverges in its focus on evaluating multiple state-of-the-art models across diverse artistic styles and historical periods. This allows for a broader assessment of model performance. 

One major area of focus has been figuring out how to evaluate and detect AI-generated images. For instance, studies like CIFAKE by Bird and Lotfi \cite{cifake} and GenImage by Zhu et al. \cite{genimage} have worked on measuring how realistic synthetic images are and developing techniques to tell them apart form human-made art. Similarly, Li et al. \cite{AIart} explored the world of adversarial AI-generated art, shedding light on the challenges of authentication and detection. These efforts are vital for assessing the authenticity of generated works, particularly in contexts where human perception plays a critical role. 

To support this kind of research, several large scale datasets have been created.
\begin{itemize}
    \item \textbf{ArtiFact Dataset:}\cite{artifact_dataset} This is a diverse mix of real and synthetic images, covering everything from human faces to animals to landscapes, vehicles and artworks. It includes images synthesized by 25 different methods, including 13 GAN-based models and 7 diffusion models.
    \item \textbf{WildFake Dataset:}\cite{wildfake_dataset} A dataset designed to assess the generalizability of AI-generated image detection models. It contains fake images sourced from the open-source community, covering various styles and synthesis methods.
    \item \textbf{TWIGMA Dataset:}\cite{twigma_dataset} A large-scale collection of AI-generated images scraped from Twitter, from 2021 to 2023, including metadata such as tweet text, engagement metrics and associated hashtags.
\end{itemize}
While these studies focus on detecting AI-generated images we focus on examining how convincingly these images replicate human-created art. Through public perception surveys, we assess whether generated paintings can be mistaken for human artwork providing insights into the models' ability to deceive the viewer aesthetically rather than algorithmically. 

Beyond detection, generated images are increasingly used as data sources for synthetic training and research applications. The work of Yang et al. \cite{aigen}
discusses the implications of using AI-generated images for training machine learning models. They explore the potential of synthetic datasets to enhance machine learning capabilities while also addressing concerns related to biases, authenticity, and ethical challenges.

Another direction in the field is the use of diffusion models for artistic style transfer. Researchers such as Chung et al. \cite{Style_injection} and Zhang et al. \cite{Step-aware} \cite{ArtBank} have introduced training-free methods and pre-trained diffusion models specifically designed for style adaptation. These works highlight the effectiveness of modern diffusion-based architectures in achieving high-fidelity artistic synthesis while maintaining flexibility for style injection. Furthermore, the work of Png et al. \cite{FeaST} proposes a feature-guided approach that improves control over the stylistic aspects of the generated output. 

The creative applications of generative AI has also been widely discussed. Haase et al. \cite{inspiring-Haase} explore the role of generated imagery in inspiring human creativity, particularly in design workflows. Similarly Barros and Ai \cite{designing-barros} investigate the integration of text-to-image models in industrial design, while Vartiainen and Tedre \cite{craft-edu} examine their use in craft education. We complement these works by examining the limitations of generative tools in artistic fidelity, particularly their struggle with maintaining compositional balance, avoiding anachronisms, and ensuring stylistic coherence. We highlight critical shortcomings such as overuse of hyperrealism, anatomical distortions and misinterpretations of historical context, which could be key obstacles to seamless integration into professional artistic workflow.  

Furthermore, a growing body of work focuses on understanding the emergent capabilities of large language models and their application in aesthetic evaluation. Studies such as those by Wei et al. \cite{emergent} and Du et al. \cite{loss_perspective} discuss how LLMs develop new abilities, such as the preference for certain artistic styles. Wang et al. \cite{evaluation-metrics} analyze evaluation metrics for generative images, offering insights on how to assess AI-generated art both quantitatively and qualitatively. These studies can be expanded with our proposed dataset, which unlike other existing ones is a controlled dataset of synthetic artworks. 

\section{Methodology}
\label{sec:methodology}
In this section we outline our methodology for the creation of
the dataset, the selection of models, and their evaluation.

\subsection{Creation of the dataset, aims, methodology used for data acquisition}

The most delicate point in the creation of the dataset was
the definition of the prompts. The importance of providing well-structured prompts for style-transfer operations is well known,
due to their direct impact on the quality and relevance of the generated outputs \cite{prompts-Oppenlaendr,prompts-Sanchez23,optimizing-prompts}. A clear and well-defined prompt eliminates ambiguity, ensuring that the model has a precise understanding of the desired style and content. Without this clarity, models can produce inconsistent or irrelevant results, making it difficult to achieve the intended artistic effect. 

In our case, prompts were generated with the assistance of ChatGPT, iteratively fine-tuning its output until acceptable results were obtained across different generators. A sample was deemed ``acceptable" if ChatGPT could recognize the required style in the generated image based on the prompt. 
Once finalized, the same prompt was passed to all models, and the results were compiled into a database accompanied by a rich set of metadata (see Section~\ref{sec:dataset}).

All prompts followed a common structure. They typically began with an indication of the style and historical period to imitate, sometimes reinforced by referencing a specific painter. This was followed by a detailed description of the subject, including suggestions for colors and tones. Finally, each prompt concluded with a hint about the overall sentiment or emotion the artwork was intended to convey. 

Here are a couple of examples:
\begin{itemize}
\item ``Generate a detailed winter landscape painting in the Flemish renaissance style of the second half of the XVI century. Depict a snow-covered village with small, rustic houses nestled into a hilly landscape. Include bare, slender trees in the foreground with hunters walking through the snow, accompanied by dogs. The scene should feature frozen lakes or ponds in the background, where villagers are skating and engaging in winter activities. The sky is a muted, wintry blue-gray, and the overall tone of the painting should evoke a peaceful, yet somewhat melancholic atmosphere, with intricate details showing rural life during winter."
\item ``Generate a view of Venice in the vedutism style of the first half of the XVIII century, focusing on a scene along the Grand Canal. The composition features detailed classical architecture with grand domes and facades, and gondolas moving along the canal. Add soft clouds to the sky and ensure there is little fading in the horizon, providing clear visibility of distant buildings. The color palette should include very soft blues and warm earth tones, avoiding saturated colors. The atmosphere remains calm and luminous, with minimal light-and-shadow effects, capturing the beauty and grandeur of Venice from a broad perspective."
\end{itemize}

\subsection{Models}
\label{sec:models}

Image generative models are a class of machine learning algorithms designed to synthesize novel images by learning the underlying patterns in existing data. By approximating the underlying distribution of visual data, these models generate outputs that form the foundation of various creative AI applications.

Within the domain of image generation, models are broadly categorized into Text-to-Image (Text2Img) and Image-to-Image (Img2Img) frameworks~\cite{instruct_pix2pix_i2i, img2img_turbo_i2i}, although hybrid and specialized approaches also exist. Text2Img models generate entirely new images based on textual descriptions, effectively translating linguistic cues into visual representations. In contrast, Img2Img models modify or enhance existing images by leveraging an input image as a reference while applying stylistic or contextual transformations. This study primarily focuses on Text2Img models due to their ability to create images purely from descriptive text prompts, making them particularly suited for analysing artistic style recreation.

To systematically evaluate the artistic fidelity and limitations of state-of-the-art (SOTA) commercial generative models, we selected 12 diffusion-based models, which are among the most widely used and highly regarded in the field. These models were identified based on their popularity and performance, as detailed in the Introduction. The selection was motivated by three key considerations:
\begin{enumerate}
    \item Benchmarking Established Models: Using well-established models enables the creation of a high-quality AI-generated art dataset, which could serve as a valuable resource for future research.
    \item Avoiding Training and Fine-Tuning Biases: Training a model from scratch or fine-tuning an existing open-source model would not provide a fair assessment of the out-of-the-box capabilities of these models. Our goal was to evaluate their pre-trained performance rather than their adaptability to new training objectives.
    \item Computational Constraints: Training or fine-tuning diffusion models is highly resource-intensive. Proprietary models, in particular, are trained on vast datasets with ongoing refinements by dedicated research teams, making them the most suitable candidates for assessing the current peak capabilities of image generative AI.
\end{enumerate}

Initially, 15 diffusion models were considered. 
Each model was tested using three standardized prompts to evaluate its ability to generate visually coherent and stylistically accurate images. Five researchers independently assessed the outputs based on realism, artifact minimization, and adherence to the prompt. A model was discarded if all five unanimously agreed it failed to meet these criteria.
For example, DeepFloyd IF \cite{stabilityai2023} was among the initial 15 models considered but was excluded from further experimentation. Its generated outputs frequently failed to align with the described artistic movements, particularly struggling with facial features and even simple object shapes (e.g., dogs and other animals).

The final selection of 12 models used in our study is listed in the Introduction, with key specifications summarized in Table 1. It is important to note that many of these models are proprietary, and as a result, their architectural details and training methodologies remain undisclosed.

\begin{table}[h]
\label{tab:models}
\centering
\resizebox{\columnwidth}{!}{%
\begin{tabular}{|p{0.20\textwidth}|p{0.14\textwidth}|p{0.28\textwidth}|p{0.16\textwidth}|p{0.13\textwidth}|p{0.17\textwidth}|}
\hline
\textbf{Model}             & \textbf{Creator}    & \textbf{Architecture}   &
\textbf{Conditioning}   
& \textbf{Resolution (default)}                                        & \textbf{Configurable output shape} \\ \hline

\href{Ideogram 2.0}{https://ideogram.ai/login} 
    & Ideogram AI 
    & Diffusion based architecture, not fully disclosed.
    & text-to-image 
    & 1024x1024 
    & Yes \\ \hline

\href{Flux 1.1 Pro}{https://blackforestlabs.ai/1-1-pro/}              
    & Black Forest Labs          
    & 
    Rectified FLow Transformer \cite{peebles2023scalablediffusionmodelstransformers,esser2024scalingrectifiedflowtransformers}       
    & text-to-image                                                         & 2048x2048 
    & Yes \\ \hline
    
\href{Dall-E 3}{https://openai.com/index/dall-e-3/} (via ChatGPT-4o) 
    & \href{OpenAI}{https://openai.com/research/}              
    & Diffusion based architecture, not fully disclosed.
    & text-to-image                                              
    & 1024x1024                                                        
    & No  \\ \hline
    
\href{Firefly Image 3}{https://firefly.adobe.com/}           
    &  Adobe               
    &  Diffusion based architecture, not fully disclosed.\cite{Adobe}
    & text-to-image, image-to-image                                        
    & 2048x2048         
    &  Yes \\ \hline
    
\href{OmniGen}{https://github.com/VectorSpaceLab/OmniGen}
    & Beijing Academy 
    & Latent Diffusion Model \cite{omnigen_t2i}.
    & multimodal-to-image
    & 2048x2048 
    & Yes \\ \hline
\href{Leonardo Phoenix}{https://leonardo.ai}          
    & Leonardo Interactive Pty          
    & Diffusion based architecture, not fully disclosed.
    & test-to-image, image-to-image                                         & 1024x1024                                                         
    & Yes \\ \hline
\href{Midjourney V6.1}{https://www.midjourney.com/imagine}
    & Midjourney          
    & Diffusion + Transformeers framework, not fully disclosed.                    & text-to-image, image-to-image                                         & 1024x1024      
    & Yes \\ \hline
\href{Stable Diffusion 1.5}{https://huggingface.co/stable-diffusion-v1-5/stable-diffusion-v1-5}       
    & Stability AI (dismissed)         
    & Latent Diffusion Model (LDM) \cite{stable-diffusion}
    & text-to-image, image-to-image                                         & 512x512                                                          
    & Yes \\ \hline
\href{Stable Diffusion 3.5-large}{https://stability.ai/news/introducing-stable-diffusion-3-5} 
    & Stability AI      
    & Advanced LDM framework with CLIP and T5 text encoders  \cite{stable_diffusion_3_5_large_t2i}. 
    & text-to-image, image-to-image \cite{QK_normalization} 
    & 1024x1024 
    & Yes \\ \hline
\href{Flux.1 Schnell}{https://blackforestlabs.ai/announcing-black-forest-labs/}            
    & Black Forest Labs 
    & Fast version of Flux.1.1 \cite{flux_schnell}, trained using latent adversarial diffusion distillation
    \cite{diffusion_distillation}.
    & text-to-image 
    & 2048x2048 
    & Yes \\ \hline
\href{Kolors 1.5}{https://klingai.com/text-to-image/new}  
    & Kuaishou Kolors team - Kling AI 
    & Large-scale latent diffusion based model.\cite{kolors}
    & text-to-image, image-to-image
    & 1024x1024
    & Yes \\ \hline
\href{Auto-Aesthetics v1}{https://neural.love/blog/auto-aesthetics-v1-ai-art-revolution}  
    & Neural.love 
    & Not disclosed \cite{NeuralLoveAI_2024} & text-to-image
    & 1024x1024 
    & Yes \\ \hline
\end{tabular}%
}
\caption{Comparison between the used Models.}
\label{tab:generative_models_table}
\end{table}



\subsection{Evaluation}

Models are evaluated based on two distinct and orthogonal criteria, each addressing a crucial aspect of their performance.

\begin{itemize}
\item Authenticity. The first criterion evaluates the model's ability to generate samples that are sufficiently realistic and convincing, such that they could be mistaken for artifacts created by a human. This involves assessing the quality of the generated output in terms of visual coherence, attention to detail, and overall believability. A high score in this area indicates that the model produces outputs that closely mimic human creativity and craftsmanship.

\item Adherence to Prompt Instructions. The second criterion focuses on the model's capacity to accurately follow the detailed instructions specified in the prompt. This involves assessing how well the generated outputs align with the intended artistic style, thematic elements, or any specific requirements outlined. Success in this area demonstrates the model's ability to interpret and faithfully execute complex and nuanced instructions.
\end{itemize}

These two evaluation criteria are deliberately designed to be independent. While a model may excel in producing outputs faithfully mimicking human art crafts, it might still fail to accurately adhere to the stylistic constraints of the prompt, or vice versa. By assessing these dimensions separately, we aim to obtain a comprehensive understanding of the model's strengths and weaknesses across both realism and prompt alignment.

The way we addressed these criteria in our surveys will be 
described in Section~\ref{sec:surveys}.

\section{The AI-pastiche Dataset}
\label{sec:dataset}

AI-Pastiche is a carefully curated dataset comprising 953 AI-generated paintings in famous artistic styles. These images were produced using manually crafted text prompts and include comprehensive metadata describing their generation details. The dataset was created using 73 carefully crafted prompts, with over 20 images generated per prompt across the selected generative models (Section \ref{sec:models}). From this pool, the highest-quality images were manually selected, ensuring fidelity to artistic styles and overall visual appeal.

\subsection{Dataset Objectives}
The two primary purposes of the dataset are:

\begin{enumerate}
\item Analyzing the capabilities and limitations of SOTA generative models in accurately recreating well-known painting styles.
\item Providing a high-quality AI-generated painting dataset for the research community, facilitating future studies on generative AI in artistic domains.
\end{enumerate}

While the current dataset consists of 953 carefully selected images, we plan to expand it in future iterations, incorporating additional artistic styles and more diverse prompts to further evaluate model performance and limitations.

\subsection{Metadata and Composition}
\label{sec:metadata}
The AI-Pastiche dataset includes detailed metadata for each generated painting, summarized in Table 2. It is important to note that attributes such as subject, style, and period correspond to the intended description in the prompt rather than a direct analysis of the generated image itself.

At present, the dataset exhibits some stylistic imbalances, particularly in terms of artistic periods and movements. As shown in Table 3, the majority of paintings emulate XIX-th and the XX-th century styles, with Renaissance, Impressionism, Romanticism, and Baroque being the most represented artistic movements. In future expansions, we aim to mitigate these imbalances by incorporating a broader range of historical styles and more diverse prompts.

Upon publication of this study, the AI-Pastiche dataset will be made publicly available on Kaggle to facilitate open research and further exploration of generative AI in artistic applications.

\begin{table}[h]
\label{tab:metadata}
    \centering
    \renewcommand{\arraystretch}{1.3} 
    \begin{tabularx}{\textwidth}{|>{\centering\arraybackslash}m{3.5cm}|X|}
        \hline
        \textbf{Attribute} & \textbf{Description} \\ 
        \hline
        \textbf{Generative Model} & The model used to generate the image. The list of models is provided in Section \ref{sec:models}. \\ 
        \hline
        \textbf{Subject} & A list of tags describing the image content based on the prompt. There are approximately 50 different tags, including categories such as “landscape,” “animals,” and “trees.” Some tags also represent colour schemes or tonal impressions, such as “gold,” “soft tones,” and “vibrant tones.” Tags are stored as a comma-separated list. \\ 
        \hline
        \textbf{Style} & A synthetic label describing the artistic style of the image. Styles include Renaissance, Baroque, Rococo, Classicism, Romanticism, Realism, Satirical, Impressionism, Art Nouveau, Naïve, Expressionism, Futurism, Cubism, Dadaism, Fauvism, Abstractionism, Symbolism, and Surrealism. \\ 
        \hline
        \textbf{Period} & The historical period or century to which the intended painting style belongs, as specified in the prompt (e.g., 18th century, Renaissance). \\ 
        \hline
        \textbf{Prompt} & The full text prompt used to generate the image. \\ 
        \hline
        \textbf{Generated Image} & The identifier of the image. \\ 
        \hline
    \end{tabularx}
    \caption{Description of the attributes used in the dataset.}
    \label{tab:attributes}
\end{table}

\begin{table}[h]
    \centering
    \renewcommand{\arraystretch}{1.2} 
    \makebox[\textwidth]{ 
     \begin{minipage}{0.45\textwidth}
            \centering
            \small
        \begin{tabular}{lcc}
            \toprule
            \textbf{Period} & \textbf{Total} & \textbf{\%} \\
            \midrule
            XX century & 307 & 32.2 \\
            XIX century & 289 & 30.3 \\
            XVI century & 153 & 16.1 \\
            XVII century & 117 & 12.3 \\
            XV century & 49 & 5.1 \\	
            XVIII century & 38 & 4 \\	
            \bottomrule
        \end{tabular}
        \caption{AI-pastiche statistics by Period.}
        \label{tab:composition_by_period}
        \end{minipage}
       \hfill
        \begin{minipage}{0.45\textwidth}
            \centering
            \small
            \begin{tabular}{lcc}
                \toprule
                \textbf{Style} & \textbf{Total} & \textbf{\%} \\
                \midrule
                Renaissance & 202 & 21.2 \\
                Impressionism & 136 & 14.3 \\
                Romanticism & 92 & 9.7 \\
                Baroque & 86 & 9.0 \\
                Realism & 60 & 6.3 \\
                Surrealism & 50 & 5.2 \\
                Dadaism & 44 & 4.6 \\
                \bottomrule
            \end{tabular}
            \caption{Most represented Styles.}
        \end{minipage}
        }
\end{table}

\section{The surveys}
\label{sec:surveys}
In order to evaluate the performance of the models along the criteria discussed in the methodology section \ref{sec:methodology}. we implemented and collected data from two distinct surveys.

\subsection{Authenticity}
With authenticity, we refer to the extent to which a model generates outputs that convincingly resemble human-made creations.

The evaluation was conducted using a survey-based approach, where participants were asked to classify images as either AI-generated or human-made. For the human-made paintings, we used a subset of open-access images from the \href{National_Gallery_of_Art}{https://www.nga.gov/open-access-images.html} in Washington. Participants were shown a set of 20 images, one at a time and in sequence, comprising a random mix of genuine and AI-generated works, and were asked to classify each image individually.

To ensure unbiased and reliable responses, the survey presented the images in randomized order, without any metadata or contextual information that could hint at their origin. This design encouraged participants to base their judgments solely on the visual and stylistic qualities of the images.

The survey was conducted anonymously. For privacy reasons, no personal information was collected. However, given the focus on European painting, we recognized that cultural background might influence participants' perceptions. To assess this, we asked participants whether they identified with a European cultural background, with the option to decline to answer.

The survey reached approximately 600 participants, selected from a diverse pool to capture a wide range of perspectives. Most of the  participants were students and colleagues, suggesting a relatively high level of education and some familiarity with artistic aesthetics, though typically without formal training in art critique. This selection was intentional, as it reflected the anticipated audience for AI-generated art in real-world scenarios. The study aimed to evaluate the perceptual authenticity as it might be experienced by the general public.

The survey is still accessible at the following page: \href{AI-pastiche_Survey}{https://script.google.com/macros/s/AKfycbzEnn5jTRW5sU7U98h8nyz6ufyk0uJjhAeOhe2LlbFk7PfpKVmtUuHFKVyyLwAOwIwl/exec}.

 \subsubsection{Adherence to Prompt Instructions}
The purpose of this evaluation is to assess each generated image based on its alignment with the requirements specified in the given prompt.

This classification task is significantly more complex than the previous one, as it requires a careful reading and thorough understanding of the prompt, as well as a comparative evaluation of outputs from different models. 
For this reason, we decided to limit participation to a selected number
of members, comprising people of our research group, colleagues of
the department of fine arts, and some of their students. While the
collective number of participants was sensibly smaller than for the first survey, each person
evaluated multiple prompts, resulting in 5706 entries with an average of about 475 assessments for each model.

We also considered the possibility of conducting a fully automated evaluation using techniques like CLIP \cite{CLIP} or similar models. However, the ability of such embeddings to accurately capture nuanced factors—such as artistic style, historical period, or other subtle attributes—remains uncertain. The data collected through this survey can also serve to clarify this issue.
This relates to our broader research program focused on investigating the aesthetic capabilities of Large Language Models, and we plan to address CLIP, among other models, in our future research.

Our evaluation metric is based on the subjective assessment of how well each image reflects the requirements of the prompt. While it is theoretically possible to rank the generated images along a continuous scale, the inherent complexity of the task and the subjective nature of the evaluations led us to simplify the process. Instead, images are categorized into three broad classes: low, medium, and high alignment with the prompt.

These classifications —low, medium, and high— are not absolute or universal but are defined relative to the specific set of images generated for each prompt. This relative approach ensures that the evaluation accounts for the context and inherent variability within each batch of images.


\section{Results}
\label{sec:results} 
In this section, we report and analyze the results of our survey. It is important to emphasize that our goal is not to compare the performance of different models, but
rather to provide a clearer understanding of the current state of the field. Our focus is on identifying the persistent challenges faced by generative models, highlighting specific problem areas, and discussing potential directions for improvement. By examining these limitations, we aim to contribute to the broader discourse on how these models can be refined and enhanced for more reliable and aesthetically convincing outputs.

\subsection{Authenticity}
\begin{figure}[h]
    \centering
    \includegraphics[width=0.5\linewidth]{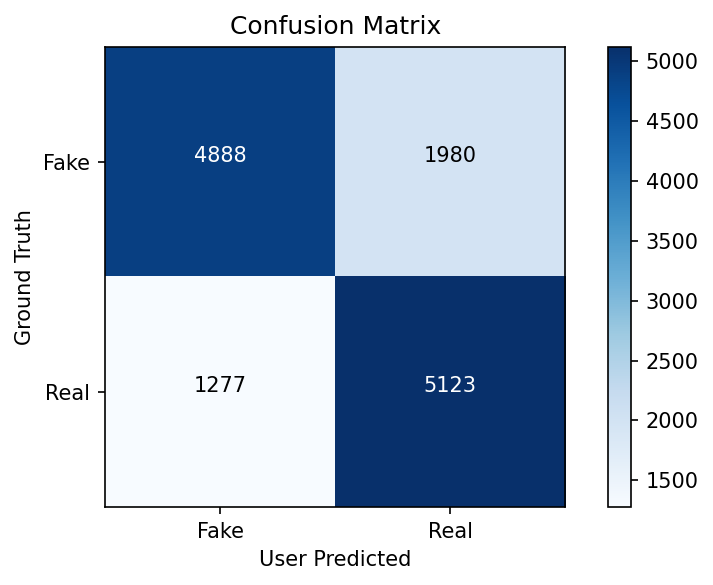}
    \caption{Confusion matrix}
    \label{fig:confusion}
\end{figure}

In Figure \ref{fig:confusion}, we show the confusion matrix
resulting from the survey: overall, around 28\% of AI-generated images were mistakenly attributed to Humans.
Interestingly enough, a slighly lower but still relevant number of Human-generated images were attributed to AI: in this case, the misclassification percentage is around 20\%.

\begin{figure}[h]
    \centering
    \includegraphics[width=0.9\linewidth]{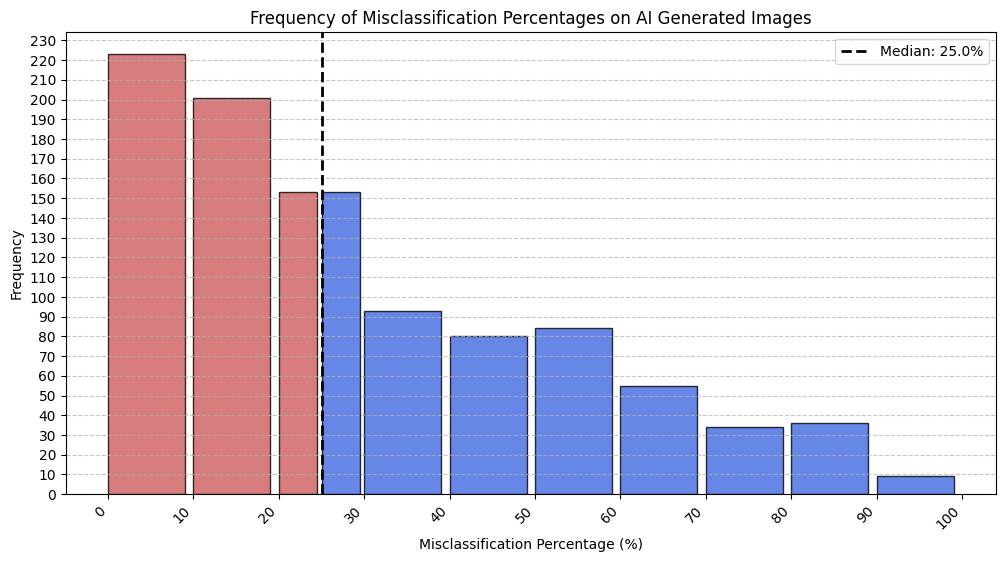}
    \caption{Distribution of Misclassification Percentages on AI-Generated Images}
    \label{fig:distribution_misclassification_on_images}
\end{figure}
Figure \ref{fig:distribution_misclassification_on_images} illustrates the frequency distribution of misclassification percentages for AI-generated images in the dataset. The distribution is skewed toward lower misclassification percentages, with a small subset of images achieving a perfect authenticity score.

Since it is not in the purpose of this work to make a
ranking of models, but merely to understand the overall
state-of-the-art, we only provide a syntetic evaluation
for the six best models, as evinced from  our survey. 
The results are summarized in Table \ref{tab:authenticity}.

\begin{table}[h]
    \centering
    \begin{tabular}{|c|c|c|c|}\hline
         {\bf model} & {\bf total count} & {\bf misclassified} & {\bf ratio}\\\hline
        Ideogram & 532  & 263  & 0.49\\\hline
        Midjourney  & 598  & 257 & 0.43 \\\hline
        Stable-diffusion-3.5-large & 572 & 204 & 0.36 \\\hline
        Stable Diffusion 1.5 & 600 & 195 & 0.33 \\\hline 
        OmniGen & 642 & 206 & 0.32 \\\hline
        Dall$\cdot$E 3 & 562 & 169 & 0.30 \\\hline
        {\bf Total} & 6868  & 1980 & 0.29 \\\hline
    \end{tabular}
    \caption{Performance of Models in terms of the perceived 
    autheticity. We only list the six most perfomant models, according to our survey. The total refer to all models.}
    \label{tab:authenticity}
\end{table}
The best performing model appears to be Ideogram, achieving an impressive authenticity rate close to 50\%. It is also noteworthy that relatively older models, such as Stable Diffusion 1.5 and Omnigen, perform comparatively well against more recent competitors. As we will see from the results of the second survey (see Section \ref{sec:adherence}), this is partly due to these models adopting a more liberal interpretation of the prompt, often sacrificing strict prompt adherence in favor of aesthetic quality.

\begin{figure}[h]
    \centering
    {\footnotesize
    \begin{tabular}{ccc}
         \includegraphics[height=0.28\linewidth]{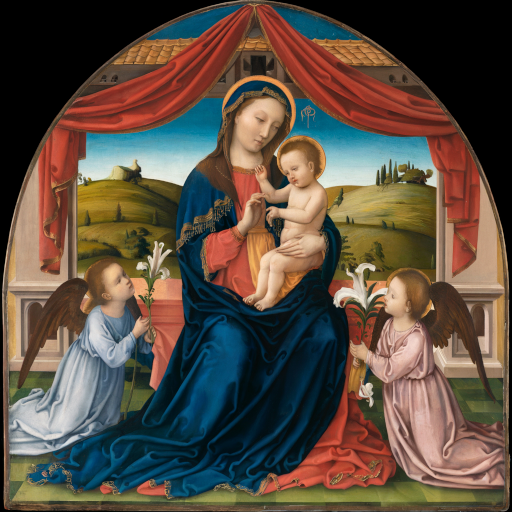} & 
          \includegraphics[height=0.28\linewidth]{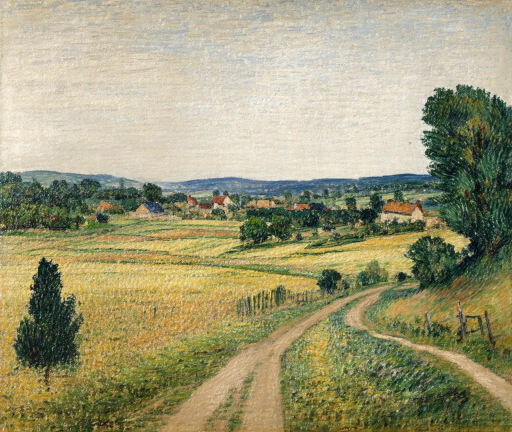} &  
          \includegraphics[height=0.28\linewidth]{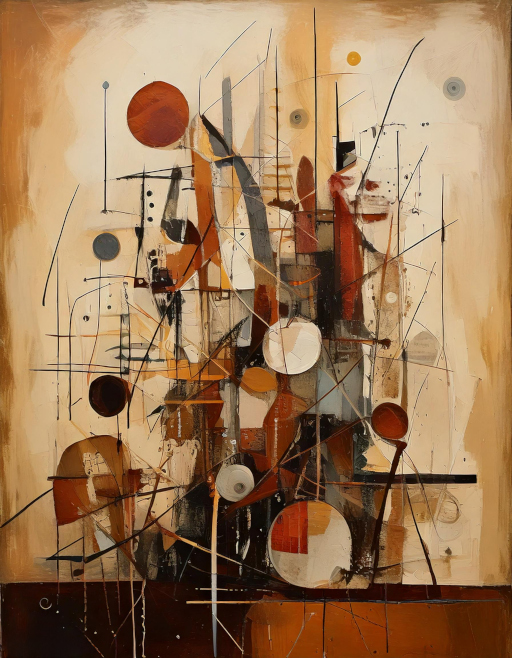}
          \\
          (a) Ideogram & (b) Stable-diffusion-3.5-large & (c) Midjourney\\
          \includegraphics[height=0.28\linewidth]{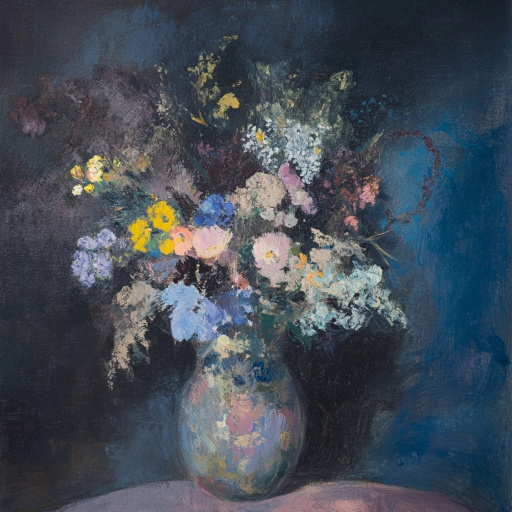} 
          &
          \includegraphics[height=0.28\linewidth]{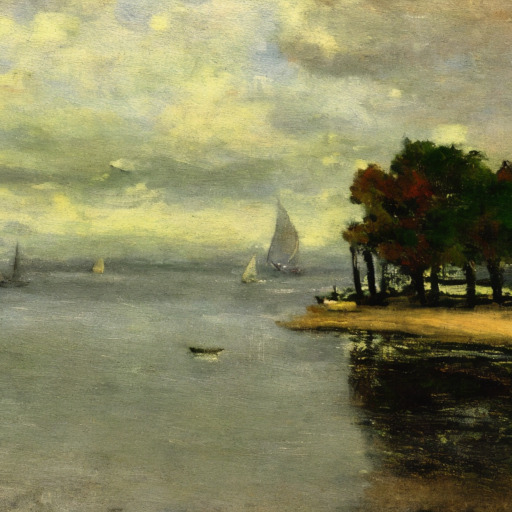} &  \includegraphics[height=0.28\linewidth]{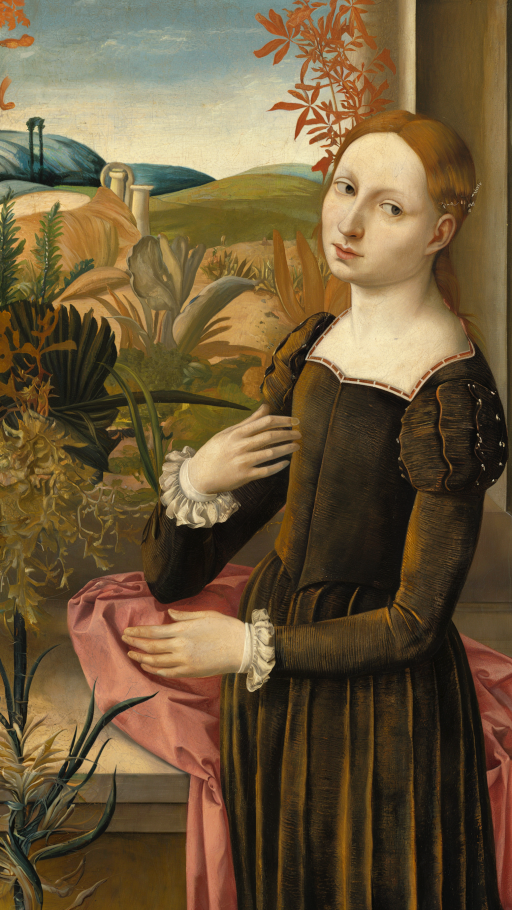} 
          \\
          (e) Firefly & (f) Stable-diffusion-1.5 & (g) Ideogram  
    \end{tabular}
    }
    \caption{Examples of convincing AI-generated examples of different Styles and Periods, according to the results of our survey.}
    \label{fig:most_convincing}
\end{figure}

Some examples of AI-generated artifacts of different styles and periods among the most frequently classified as human-made according to our survey are shown in Figure \ref{fig:most_convincing}.

A per-period investigation (see Table \ref{tab:authenticity_per_period}) show that, not surprisingly, generative
models perform particularly well in mimicking art of the
last century, and (some styles) of the XIX century. 
They clearly seem to be in much more trouble in producing convincing artifacts of previous periods.

\begin{table}[h]
    \centering
    \begin{tabular}{|c|c|c|c|}\hline
         {\bf period} & {\bf total count} & {\bf misclassified} & {\bf ratio}\\\hline
XX century     &     2095    &   697  & 0.33\\\hline
XIX century    & 1955    & 541    & 0.28\\\hline
XVII century   & 786     & 208    & 0.26 \\\hline
XV century     & 331     & 84     & 0.25 \\\hline
XVI century    & 1083 &  265 & 0.24 \\\hline
XVIII century   & 261 & 57 & 0.22 \\\hline
    \end{tabular}
    \caption{Caption}
    \label{tab:authenticity_per_period}
\end{table}

Analyzing results according to artistic styles is complicated by the current underrepresentation of certain movements. For instance, as reported in Table \ref{tab:authenticity_per_style}, the style that the models have been most comfortable with is `Art Nouveau'. 

\begin{table}[h]
    \centering
    \begin{tabular}{|c|c|c|c|}\hline
         {\bf period} & {\bf total count} & {\bf misclassified} & {\bf ratio}\\\hline
art nouveau & 104  & 49 & 0.47 \\\hline
cubism  & 232 & 92 & 0.40 \\\hline
satirical & 74 & 29 & 0.39 \\\hline
impressionism & 922 & 350 & 0.38 \\\hline
dadaism & 320 & 118 & 0.37 \\\hline
futurism & 114 & 42 & 0.37 \\\hline
classicism & 273 & 99 & 0.36 \\\hline
fauvism & 119 & 40 & 0.34 \\\hline
expressionism  & 170  & 57 & 0.34 \\\hline
symbolism  & 302  & 98 & 0.32 \\\hline
vedutism  & 92  & 26 & 0.28 \\\hline
renaissance  & 1458  & 355  & 0.24 \\\hline
romanticism  & 635  & 154  & 0.24 \\\hline
abstractionism  & 91 & 20 & 0.22 \\\hline
baroque & 574 & 123  & 0.21  \\\hline
realism & 402 & 85   & 0.21 \\\hline
surrealism & 334 & 70 & 0.21 \\\hline
rococo & 157 & 30 & 0.20 \\\hline
naive & 201 & 33 & 0.16 \\\hline
    \end{tabular}
    \caption{Caption}
    \label{tab:authenticity_per_style}
\end{table}

\begin{figure}[h]
    \centering
    {\footnotesize
    \begin{tabular}{ccc}
         \includegraphics[height=0.3\linewidth]{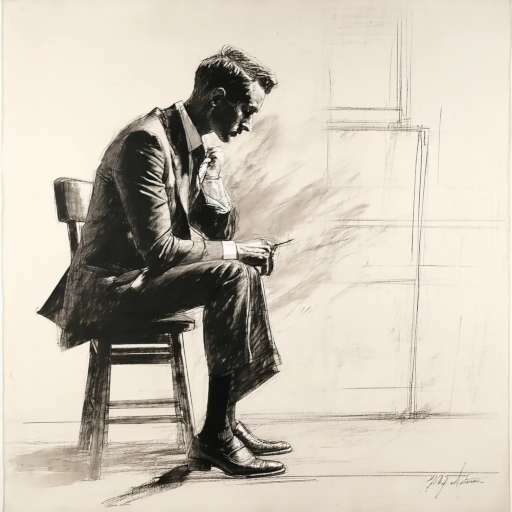} & 
          \includegraphics[height=0.3\linewidth]{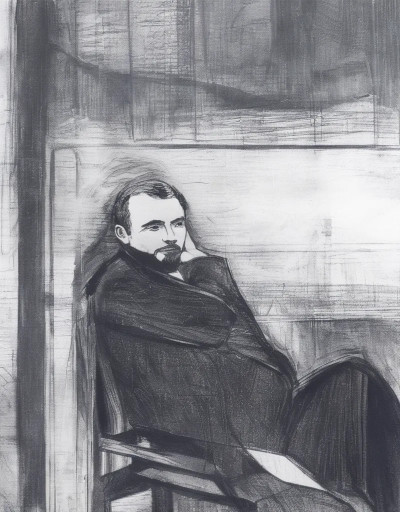} &  \includegraphics[height=0.3\linewidth]{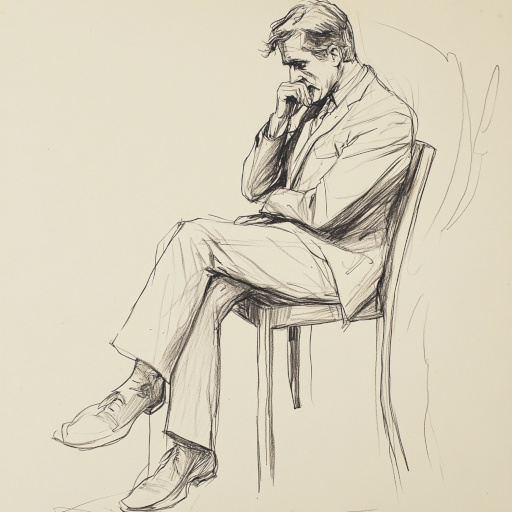} \\
          (a) Stable-diffusion-3.5-large & (b) OmniGen & (c) Midjourney 
    \end{tabular}
    }
    \caption{``Art nouveau" examples. The sketchy nature of the subject
    specified by the prompt adapted particularly well to the capacities of generative models.}
    \label{fig:art_nouveau}
\end{figure}

However, we have only a single prompt associated with this label, depicting a pencil sketch of a seated man in a pensive attitude. A few examples are shown in Figure \ref{fig:art_nouveau}. Due to the schematic simplicity of both the subject and the technique, it is not surprising that many of the AI-generated artifacts have been mistakenly perceived as human-made.

\begin{figure}[h]
    \centering
    {\footnotesize
    \begin{tabular}{ccc}
         \includegraphics[height=0.3\linewidth]{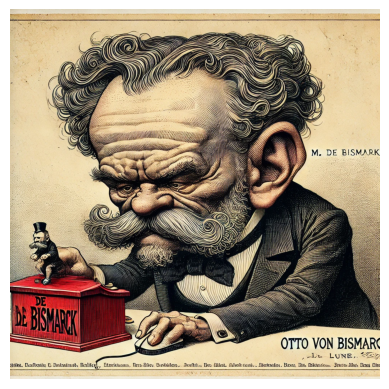} & 
          \includegraphics[height=0.3\linewidth]{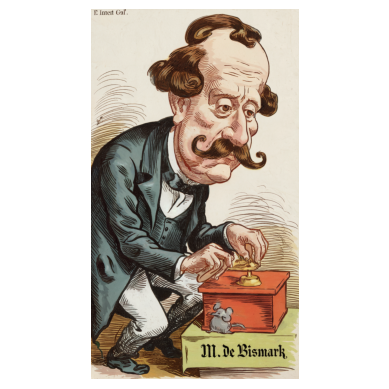} &  \includegraphics[height=0.3\linewidth]{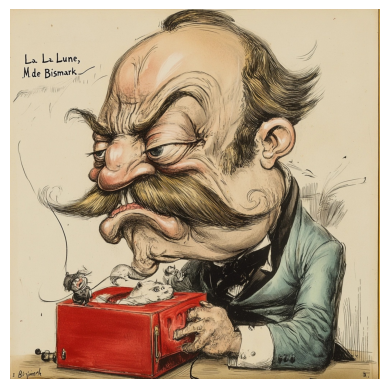} \\
          (a) Dalle$\cdot$E & (b) Ideogram & (c) Midjourney 
    \end{tabular}
    }
    \caption{Satirical examples.}
    \label{fig:satirical}
\end{figure}

A similar problem arises with the ``satirical" style. Again, we
have only one prompt relative to this category, referring to a 
caricature of Otto Von Bismark in the sytle of the satirical magazine ``La Lune", of the end of the XIX century. Many models created convincing 
artifacts, as illustrated in Figure \ref{fig:satirical}.

Apart from these cases, generative models appear to be more adept at imitating modern artistic styles, such as Impressionism, Cubism, Dadaism, Futurism, and similar movements. These styles often emphasize abstraction, bold shapes, and expressive brushwork, which align well with the strengths of generative models.

Conversely, models face greater challenges when attempting to replicate older artistic styles, such as Renaissance, Baroque, and Rococo. These styles are characterized by intricate details, realistic depictions, and complex compositions, which demand a level of precision and 
semantic interpretation that many models struggle to achieve.

Interestingly, the worst performance is observed when models attempt to imitate naïve art. One key reason for this difficulty is the challenge most models face in handling the ``flat" perspective typical of this style, as discussed in Section \ref{sec:landscapes}. Unlike classical or modern styles, naïve art often employs a lack of depth, disproportionate figures, and an intuitive rather than rule-based approach to composition. This contradicts the implicit biases of generative models, which are often trained to prioritize realism, shading, and perspective consistency.

\subsubsection{Distinction of results for cultural background}
As mentioned in the introduction, we asked participants to disclose their cultural background to assess its potential impact on the perception of European paintings.

In this regard, the collected data are highly unbalanced, with European participants outnumbering non-European participants by approximately six to one. As a result, any analysis of this factor must be approached with caution, as the sample distribution may limit the reliability of our findings.

\begin{figure}[h] 
    \centering
    \includegraphics[width=1\textwidth]{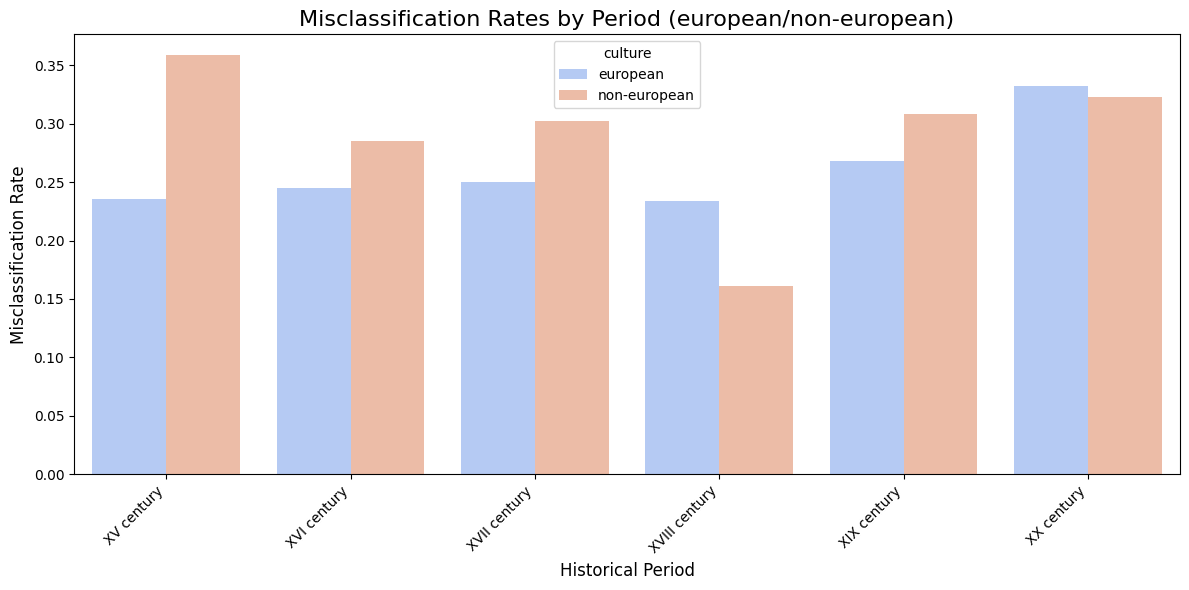} 
    \caption{Misclassification rates by periods.}
    \label{fig:barplotperiods} 
\end{figure}

The only interesting result is relative to 
the misclassification rate for different historical
periods, shown in Figure~\ref{fig:barplotperiods}.


Not surprisingly, non-European participants tend to misclassify images from the 15th, 16th, and 17th centuries more frequently, likely due to a lower level of familiarity with the artistic movements of those periods. European art from these centuries is deeply rooted in specific cultural and historical contexts, with stylistic conventions that may not be as immediately recognizable to those who have not been extensively exposed to them. 

A similar analysis across different artistic styles did not reveal any additional trends significant enough to report. 

\subsubsection{Influence of the subject}
\label{sec:subject_influence}
Our final investigation examines the influence of subject matter on the model's ability to generate artifacts that can be mistaken for human-made creations. For this analysis, we use the tags described in Section~\ref{sec:metadata}. Specifically, each prompt is represented as a multilabel binarization over its associated set of tags. We then perform a linear regression to predict the average degree of ``authenticity", as determined by the survey, for all entries associated with those tags. The analysis is restricted to tags occurring in at least two different prompts.

Naturally, we do not expect to obtain a highly accurate estimation, as additional — such as the required style and historical period — also play a role. However, our focus is not on the predicted output itself but rather on the weights assigned by the model to different tags, particularly negative tags, which may indicate categories that present challenges for the models.

After normalizing the output using Gaussian normalization, we obtain a prediction error of approximately 0.4 (compared to the unit standard deviation). As expected, the prediction accuracy is not particularly high, but it is sufficient to demonstrate a correlation between tags and perceived authenticity. 

\begin{figure}
    \centering
    \includegraphics[width=1.\linewidth]{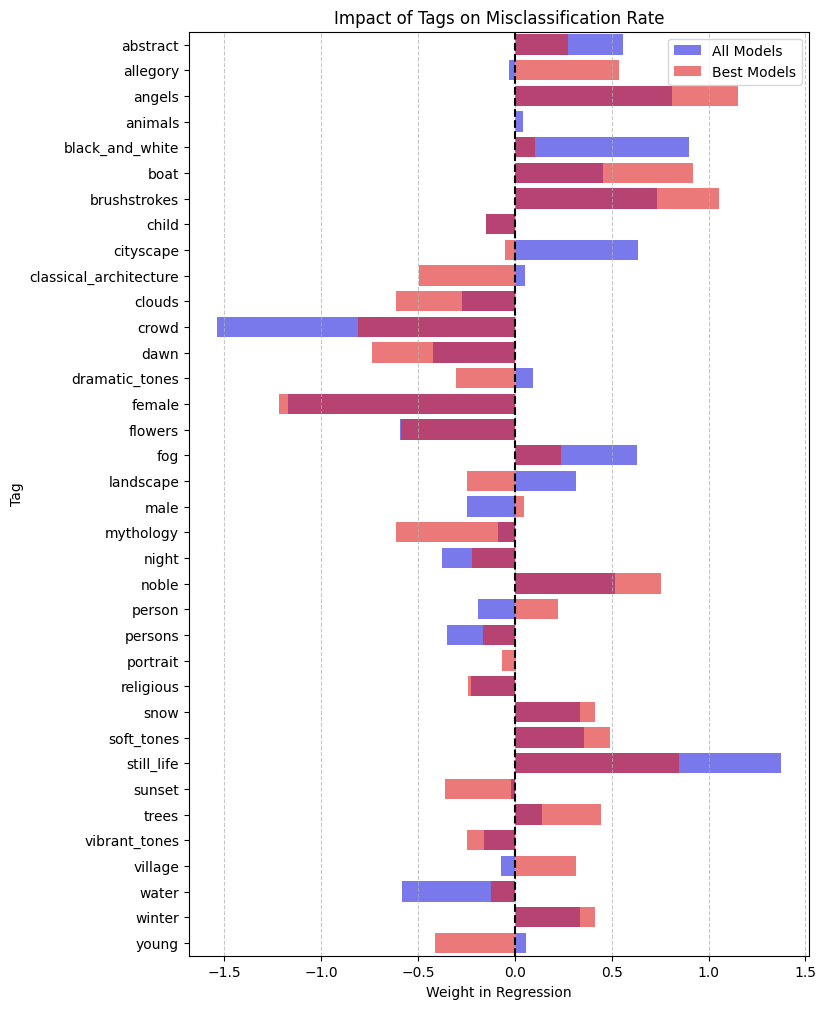}
    \caption{Impact of the tags on the perceived authenticity of the AI-generate artifact, estimated through linear regression. The analysis is restricted to tags occurring in at least two different prompts. We perform the investigation for all models (blue) and for 
    a restricted subset of models with higher authenticity 
    scores (red).}
    \label{fig:tag-weights}
\end{figure}

In Figure~\ref{fig:tag-weights} we show the weights associated with the different tags. We do the investigation for all models (blue) and for a restricted subset of models comprising Ideogram, Midjourney, Stable-Diffusion-3.5-large and Dall$\cdot$E, obtaining high scores both in authenticity and prompt adherence.

Looking at the negative scores, a notable group is composed
by tags related to humans: ``crowd", ``person", ``persons", ``child", and ``portrait". 
This provides strong evidence that generative models still struggle to represent humans convincingly when mimicking artistic painting. In addition, portraits of women tend to present more challenges compared to those of men.

This difficulty may arise from several, sometimes contrasting, factors. For example, generative models may fail to achieve realism in highly complex and dynamic scenes involving multiple people or crowds, while at the same time, they may adopt an exaggerated hyperrealism in portraits. We discuss these issues in more detail in Section \ref{sec:discussion}.

From the naturalistic point of view, ``clouds", ``flowers" and ``water" seem to have a negative impact. The tag ``flower" contrasts with ``still\_life", which, by comparison, has a significantly more positive score. In our dataset, the negative perception associated with flowers seems to be primarily linked to paintings in naïf style — one of the styles where generative models, as observed in the previous section, tend to perform the worst. 
The negative scores for ``clouds" and ``water" appear to stem from the inherent complexity of rendering these elements in a way that aligns with the stylistic and historical constraints specified in the prompt.  It is 
also interesting to observe that while the ``best" models seem to be able to cope with water in an acceptable way, their performance on ``clouds" is even worse than average.
We shall discuss this subject in more detail, in Section \ref{sec:landscapes},
where we shall also provide a few examples.

Other tags related to nature, such as ``fog", ``snow", and ``trees", do not appear to pose significant challenges for generative models. These elements are often rendered convincingly, likely due to their relatively uniform structures and the abundance of high-quality reference images available in training datasets. However, the situation changes when considering specific moments of the day. Night scenes can easily suffer from inconsistencies in lighting and contrast or from the hyperrealistic rendering of specific elements, such as the moon. More notably, dawn and sunset present particular difficulties, as generative models often struggle to capture the complex interplay of warm and cool tones, the gradual transitions in atmospheric lighting, and the way natural and artificial light sources interact during these times. These shortcomings can lead to unnatural gradients, misplaced highlights, or an overall loss of realism, making these scenarios more challenging than other elements related to nature.

The explicit request in the prompt to add visible brushstrokes frequently increases the perception of authenticity. In addition, 
models generally perform better when prompted to adopt soft, muted tones rather than vibrant or dramatic color schemes. When working with softer tones, the model is more likely to produce balanced, harmonious compositions that align well with a wide range of artistic styles. In contrast, when tasked with generating highly saturated or dramatic lighting effects, models often tend to over-interpret the request, leading to exaggerated contrasts, unnatural color blending, or an overuse of artificial-looking highlights and shadows.

Finally, models appear to struggle with subjects related to mythology and religion, due to a combination of the inherent complexity of these themes and content moderation filters that may constrain or influence their performance.

\subsection{Adherence to Prompt Instructions}
\label{sec:adherence}
This survey measured user satisfaction based on the alignment of generated images with the requirements specified in the given prompt, considering both content and style. Users rated their evaluations in three categories: ``Good", ``Medium", and ``Low." The evaluation was not intended to be absolute, but rather comparative, assessing how each model's output performed relative to others.

For instance, if a particular image was unanimously classified as "Good," this does not necessarily imply that it was a highly satisfactory interpretation of the prompt. Rather, it simply indicates that, in the collective judgment of the reviewers, it outperformed the outputs of other models.

Reviewers were encouraged to give a balanced repartition in the three categories,
to reduce the impact of the prompt complexity, and inherent variability within each batch of images.

To derive a summary score, we computed a weighted average, assigning a value of 1 to ``Good", 0 to ``Medium", and -1 to ``Low". 

\begin{figure}[h] 
    \centering
    \includegraphics[width=1\textwidth]{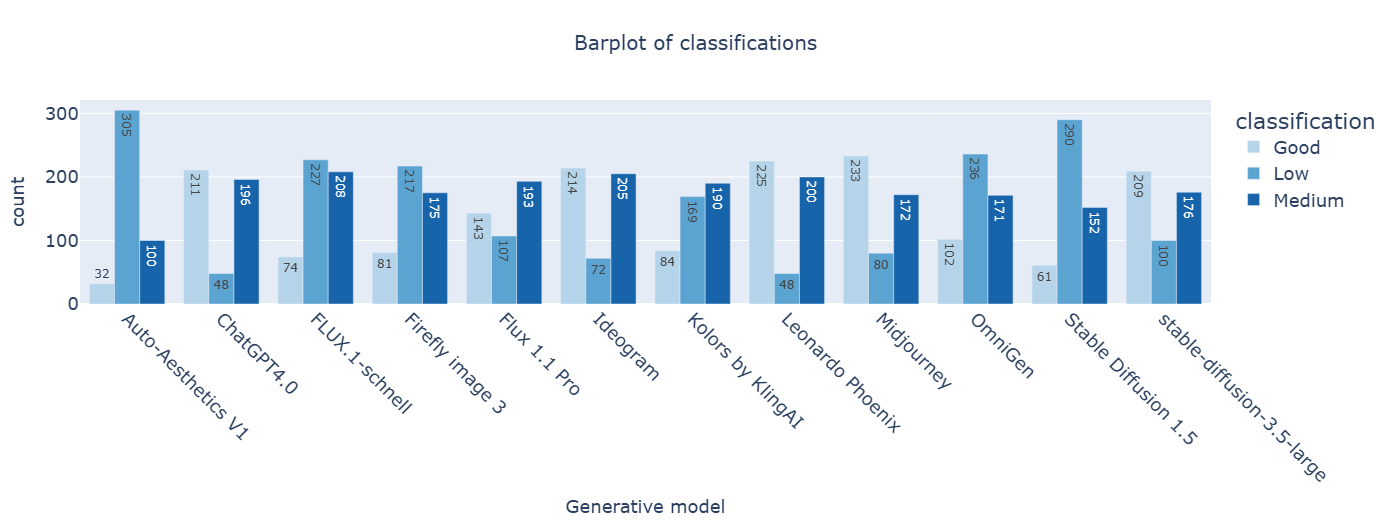} 
    \caption{Descrizione dell'immagine}
    \label{fig:plot_survey2} 
\end{figure}

The results are summarized in Figure \ref{fig:plot_survey2} and Table \ref{tab:adherence_scores}. Again, in the Table we only list the most performant models, according to out investigation.

\begin{table}[h]
    \centering
    \begin{tabular}{c|c}
    {\bf model} & {\bf average score}\\\hline\hline
  Leonardo Phoenix & 0.37 \\\hline
  Dall$\cdot$E 3 & 0.36\\\hline
  Midjourney & 0.32\\\hline
  Ideogram & 0.29 \\\hline
  stable-diffusion-3.5-large & 0.22 \\\hline
  Flux 1.1 Pro & 0.08 \\\hline
    \end{tabular}
    \caption{Average ``adherence" score. The score was computed as a weighted average of the survey results, associating value 1 to ``Good", 0 to ``Medium" and -1 to ``Low". We only list models with a better than average
    behaviour, according to our investigation.}
    \label{tab:adherence_scores}
\end{table}

It is worth noting that prompt adherence leads to a substantially different ranking compared to authenticity scores, which were evaluated without knowledge of the corresponding prompts. A model that was instructed to generate a Renaissance painting but instead produced a convincing Cubist artwork would likely receive a high authenticity score, despite failing to follow the intended artistic style.

This suggests that some models prioritize aesthetic quality over strict prompt adherence, opting for visually compelling outputs even at the expense of accuracy. This tendency is particularly evident in early-generation generative models, such as Stable Diffusion 1.5 and Omnigen, which frequently take creative liberties with prompt instructions.

Despite their loose interpretation of prompts and their occasional introduction of artifacts and distortions, these models remain among the most creative and surprising in our tests. Their ability to produce unexpected yet visually engaging results highlights a trade-off in generative AI: while newer models may achieve higher precision in style replication, earlier models often exhibit a greater degree of unpredictability and artistic exploration, which can sometimes lead to unexpectedly compelling outputs.

\section{Critical aspects of artificial generation}
\label{sec:discussion}
In this section, we highlight some of the most common and critical challenges observed in the generative models under consideration. These insights stem both from our direct experience in dataset creation and from the results of our surveys.

We structure the discussion around three major problem areas: Artifacting and Distortion (Section \ref{sec:artifacts}), Hyperrealism (Section \ref{sec:hyperrealism})
and Anachronisms (Section \ref{sec:anachronisms}).

\subsection{Artifacting and Distorsion}
\label{sec:artifacts}
One of the most evident problems is artifacting and distortion, where models fail to maintain anatomical coherence or structural integrity.
A few major instances are discussed below.

\subsubsection{Fingers, Hands and Limbs}
\begin{figure}[h]
    \centering
    {\footnotesize
    \begin{tabular}{cccc}
          \includegraphics[height=0.16\linewidth]{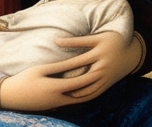} & \includegraphics[height=0.16\linewidth]{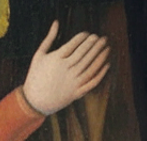} &
          \includegraphics[height=0.16\linewidth]{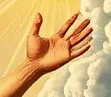} &
          \includegraphics[height=0.16\linewidth]{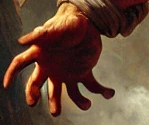} \\
          Leonardo Phoenix & Ideogram 
         & Flux1-Schnell & Dall$\cdot$E 3 \\
            \includegraphics[height=0.16\linewidth]
          {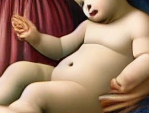} &
          \includegraphics[height=0.16\linewidth]{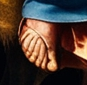} &
          \includegraphics[height=0.16\linewidth]{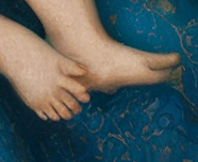} &
          \includegraphics[height=0.16\linewidth]{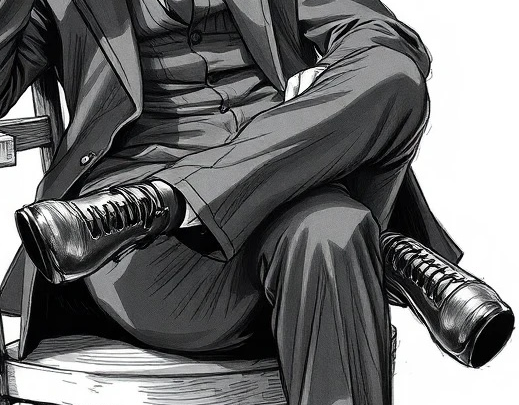} 
           \\
          Dall$\cdot$E 3 & Leonardo Phoenix & Ideogram &  Auto-Aesthetics v1  \\
           \includegraphics[height=0.16\linewidth]{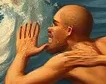} &
          \includegraphics[height=0.16\linewidth]{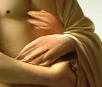} &
          \includegraphics[height=0.16\linewidth]{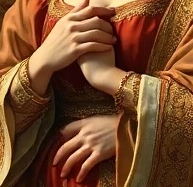} &
           \includegraphics[height=0.16\linewidth]{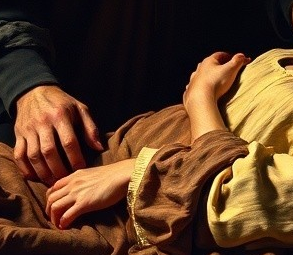} 
           \\
          Flux1-Schnell & Leonardo Phoenix & Auto-Aesthetics v1  & Auto-Aesthetics v1 
    \end{tabular}
    }
    \caption{Problems with fingers in hands and foots}
    \label{fig:fingers}
\end{figure}

Correctly rendering hands and fingers remains one of the major challenges in generative image synthesis. 
The problem is common to most of the models: some examples are given in  Figure~\ref{fig:fingers}

The problem is well known and stems from several factors. The primary challenge is that hands frequently interact with objects or other parts of the body, leading to complex occlusions and overlapping regions. This poses difficulties both during training, where the model must abstract hands and fingers from their specific context, and during generation, where the model must realistically render them within the context of these interactions.

The problem is not limited to human figures. Animals are frequently depicted with an innatural number of legs, heads, or similar distorsions.

\begin{figure}[h]
    \centering
    {\footnotesize
    \begin{tabular}{cccc}
          \includegraphics[height=0.16\linewidth]{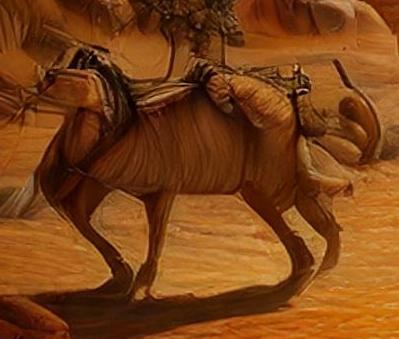} &
          \includegraphics[height=0.16\linewidth]{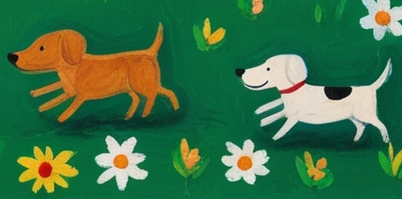} &
          \includegraphics[height=0.16\linewidth]{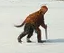} &
          \includegraphics[height=0.16\linewidth]{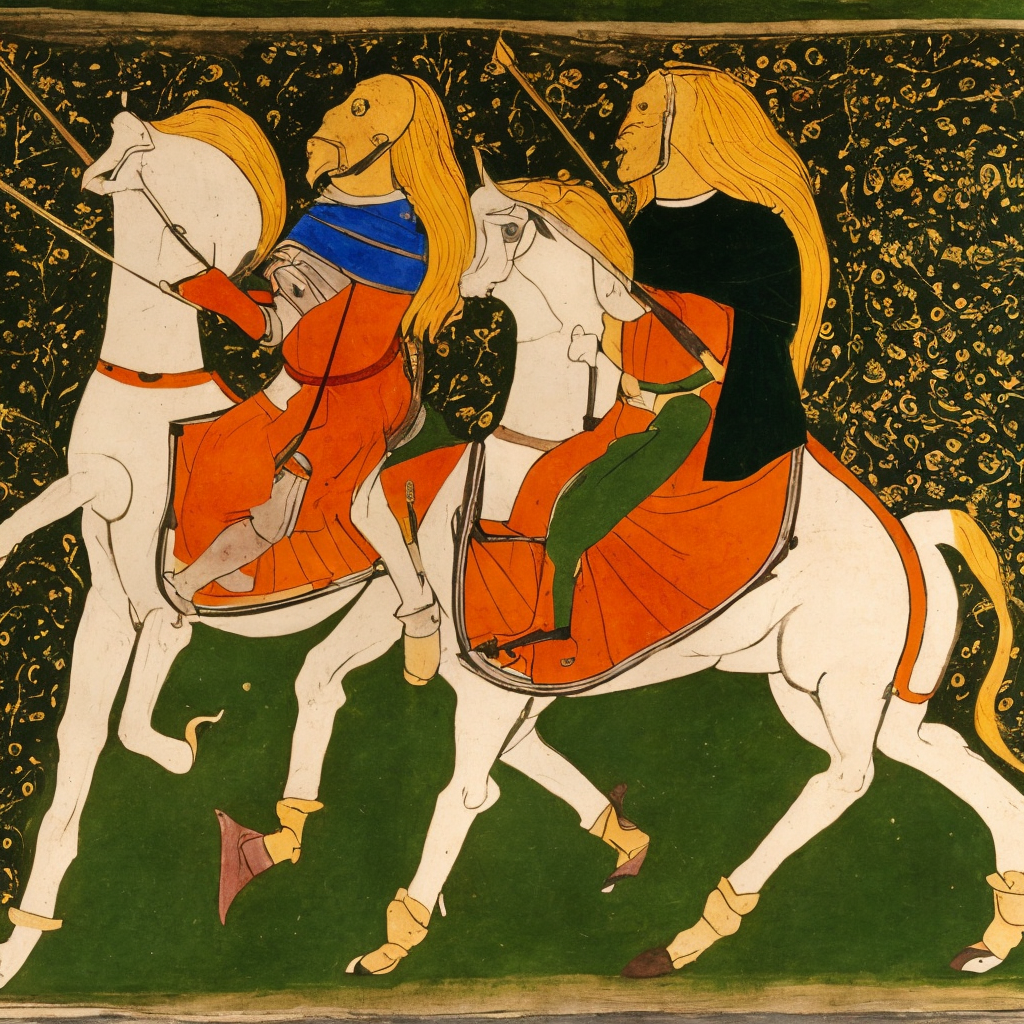} \\
          (a) Firefly & (b) Leonardo Phoenix & 
          (c) Flux1-Schnell  & (d) Stable Diff.1.5
    \end{tabular}
    }
    \caption{Caption}
    \label{fig:legs}
\end{figure}

\subsubsection{Distorsions in complex scenarios}
\begin{figure}[h]
    \centering
    {\footnotesize
    \begin{tabular}{ccc} 
          \includegraphics[height=0.24\linewidth]{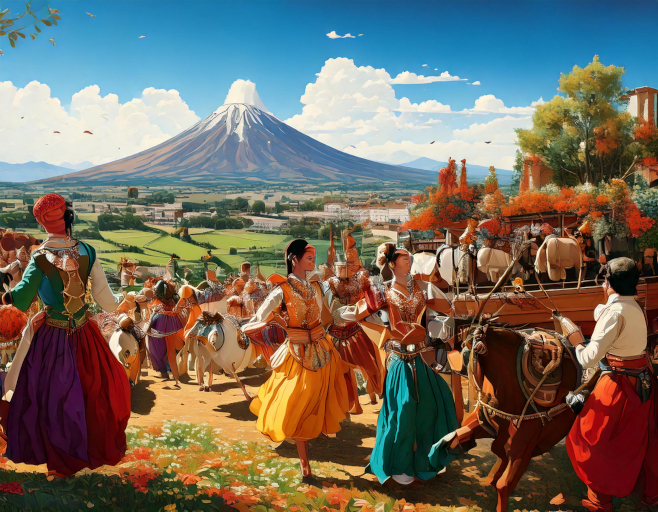} &
          \includegraphics[height=0.24\linewidth]{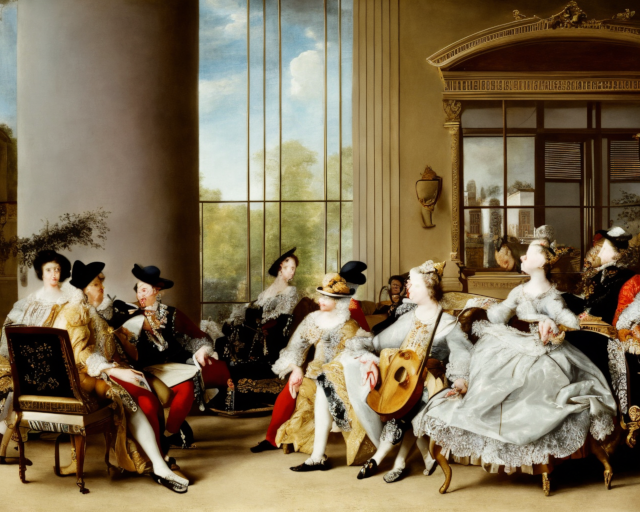} &
          \includegraphics[height=0.24\linewidth]{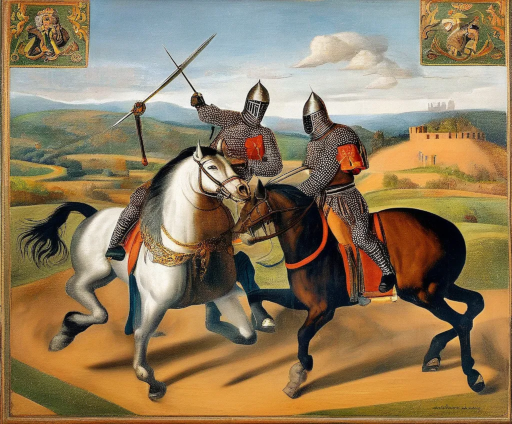} \\
          (a) Firefly & (b) Stable Diffusion 1.5 & (c) Omnigen
    \end{tabular}
    }
    \caption{Caption}
    \label{fig:dynamic}
\end{figure}

Distortions become more pronounced in complex scenarios, such as groups of people, highly dynamic scenes, or intricate architectural compositions. In these cases, maintaining a natural balance between stylistic accuracy and structural coherence remains a challenging task for most models.
A few typical examples are shown in Figure~\ref{fig:dynamic}, but nearly all models struggled with these specific prompts: (a) a traditional rural festival in the 19th-century realism style, (b) a music lesson in the Rococo style, and (c) a battle between knights in the early Renaissance style.

One frustrating limitation of current generative models is their inability to dynamically adjust generation time based on the complexity of the task. Unlike human artists, who naturally dedicate more time to intricate compositions while completing simpler ones more quickly, these models follow a fixed computational budget, regardless of the difficulty of the image being generated.

For instance, diffusion-based models operate within a predefined number of denoising steps, meaning they do not inherently ``realize" when an image requires additional refinement to resolve ambiguities in structure, perspective, or stylistic details. Whether generating a minimalist still life or a highly detailed historical battle scene, the model performs the same number of steps, often leading to overprocessing in simple cases and underdeveloped details in complex ones.


\subsection{Hyperrealism}
\label{sec:hyperrealism}
Most generative models, often optimized for photorealism, struggle to reproduce the unique nuances and distinctive qualities characteristic of artistic styles from the past. We shall discuss the issues in 
three paradigmatic cases: portraits, still lifes, and landscapes.

\subsubsection{Portraits}
\begin{figure}[h]
    \centering
    {\footnotesize
    \begin{tabular}{ccc}
          \includegraphics[height=0.29\linewidth]{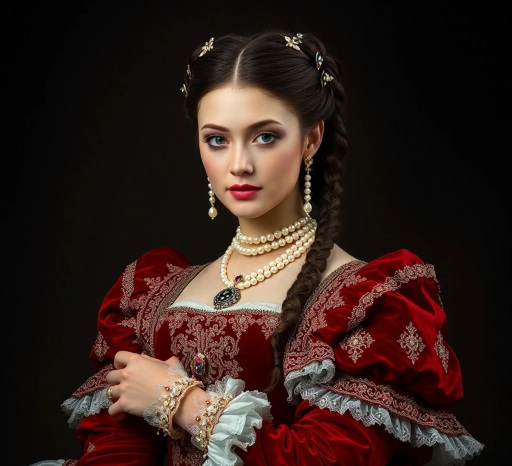} &  \includegraphics[height=0.29\linewidth]{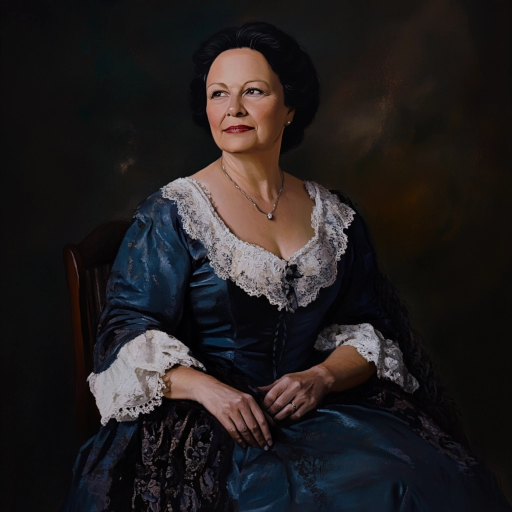} & 
          \includegraphics[height=0.29\linewidth]{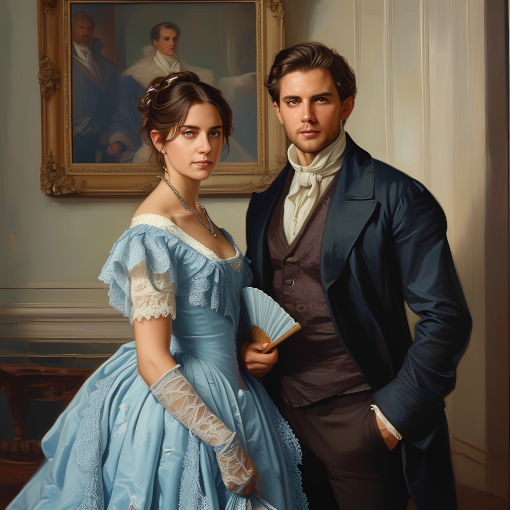}\\  (a) Auto-Aesthetics v1 & (b) Midjourney & (c) Kolors by KlingAI \\
          \includegraphics[height=0.29\linewidth]{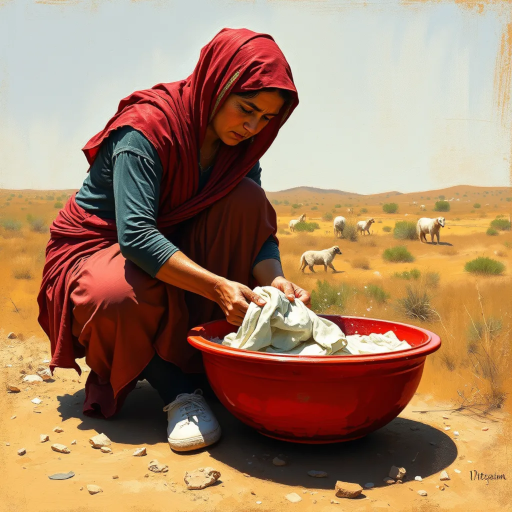} &  \includegraphics[height=0.29\linewidth]{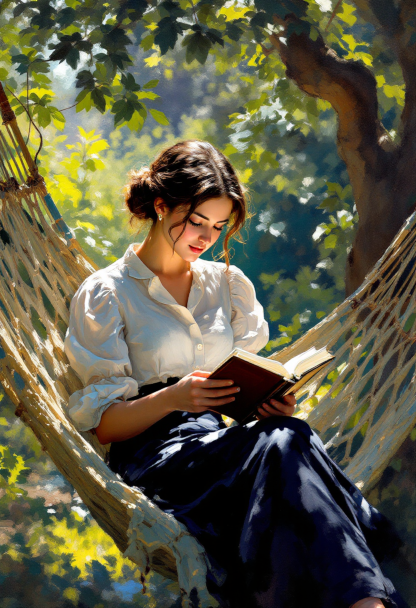} & 
          \includegraphics[height=0.29\linewidth]{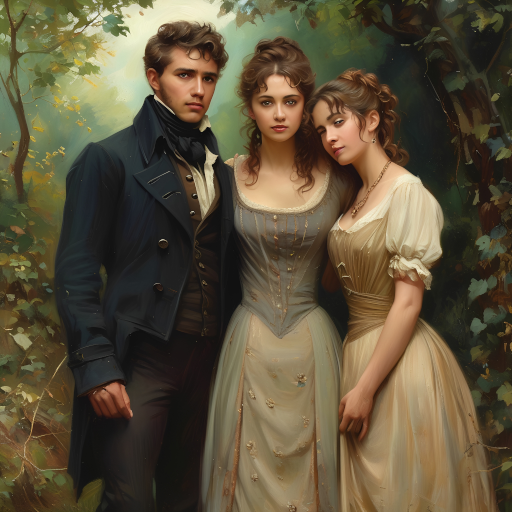}\\  
          (d) Flux 1 - Schnell & (e) Flux 1.1 - Pro & (f) Kolors by KlingAI
    \end{tabular}
    }
    \caption{Hyperrealism examples. The required styles were: (a) baroque, (b) baroque, (c) romanticism, (d) XIXth Century realism,  (e) impressionism, (f) romanticism.}
    \label{fig:hyperrealism_faces}
\end{figure}

Modern generative models often exhibit a hyperrealistic tendency when replicating facial details, often in contrast with 
the historical artistic style they were supposed to mimic according to
the prompt. This excessive sharpness and detail can create a fundamental mismatch between the expected stylistic conventions and the generated output, leading to images that feel anachronistic or unconvincing. A few examples are given in 
Figure~\ref{fig:hyperrealism_faces}

The problem becomes even more apparent when considering historical limitations in artistic materials and techniques. Painters working with oil or tempera could not achieve the pore-level skin textures or ultra-sharp reflections that modern models tend to generate by default. When a generative model introduces such hyperrealistic details into a 
Rinascimental painting or a 17th-century Dutch portrait, the output no longer aligns with the stylistic expectations of that period.

\subsubsection{Still lifes}
Another common subject where generative models struggle to restrain their tendency toward excessive realism is still life painting. While still lifes often contain highly detailed depictions of objects, traditional artistic styles—especially those from historical periods—frequently employ soft lighting, controlled textures, and a painterly touch that distinguishes them from hyperrealistic renderings.

Generative models, however, tend to overemphasize surface details, reflections, and textures, producing results that lean toward photographic realism rather than adhering to the stylistic characteristics of classical still life compositions. This issue becomes particularly noticeable in flower arrangements, fruit compositions, and table settings, where the AI-generated images may include overly sharp edges, unnatural glossiness, or exaggerated depth-of-field effects that are inconsistent with traditional oil painting techniques. 

\begin{figure}[h]
    \centering
    {\footnotesize
    \begin{tabular}{ccc}
          \includegraphics[height=0.45\linewidth]{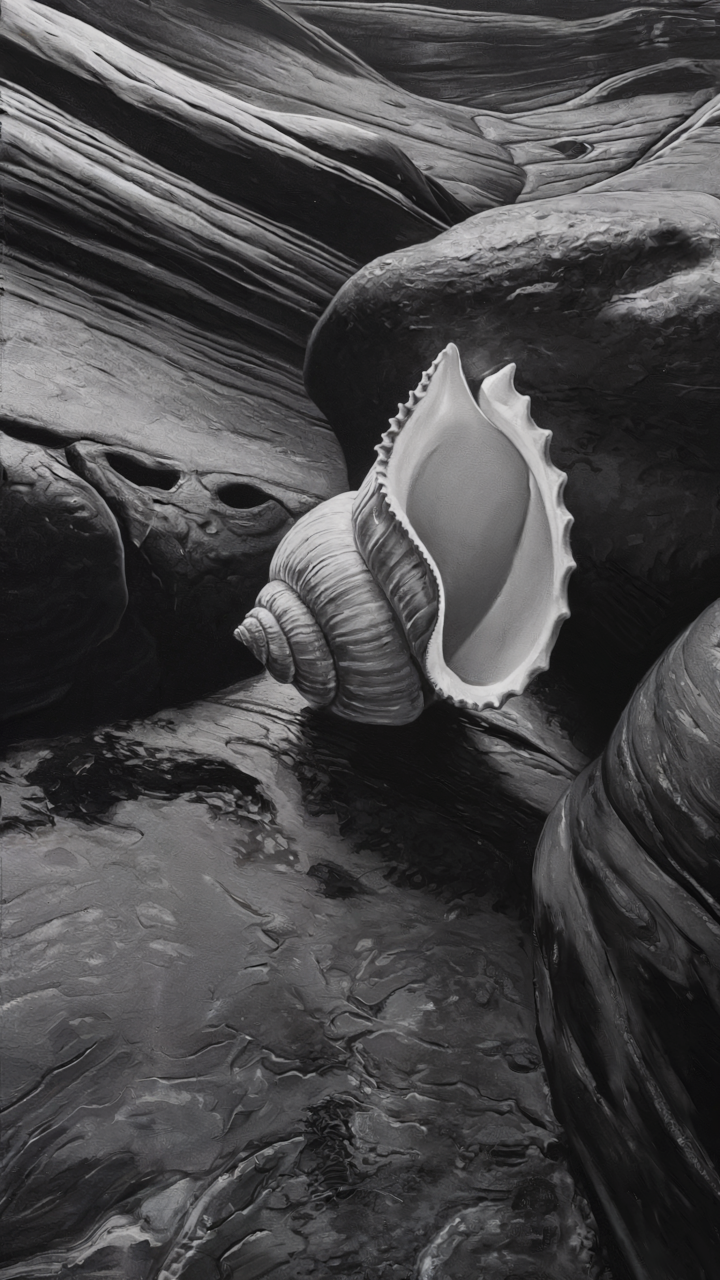} &  \includegraphics[height=0.45\linewidth]{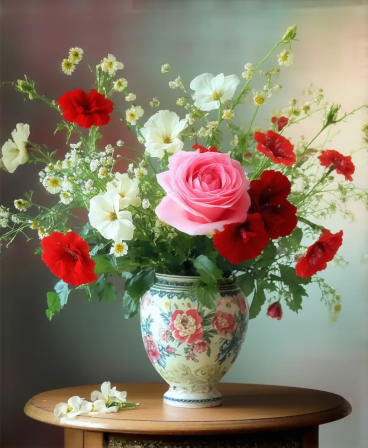} & 
          \includegraphics[height=0.45\linewidth]{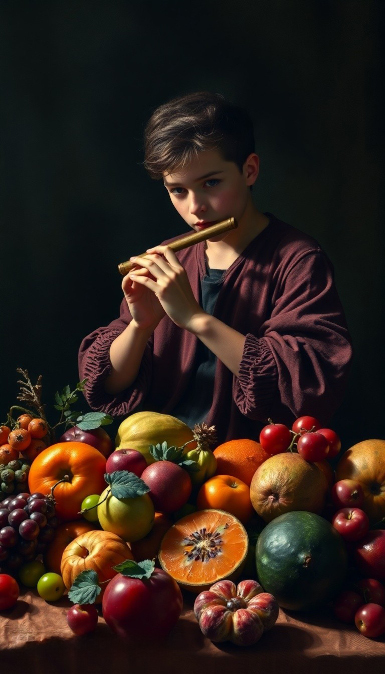}\\  
          (a) Ideogram & (b) Stable-Diffusion-3.5-large & (c) Auto-Aesthetic V1
    \end{tabular}
    }
    \caption{Hyperrealism examples. The required styles were: (a) surrealism, (b) classicism, (c) renaissance }
    \label{fig:hyperrealism_still_lifes}
\end{figure}

A few typical examples are shown in Figure~\ref{fig:hyperrealism_still_lifes}.

\subsubsection{Landscapes and cityscapes}
\label{sec:landscapes}
As evidenced by the tag analysis in Section \ref{sec:subject_influence}, the most challenging elements in the representation of naturalistic scenes are clouds and water, particularly when combined with specific times of the day—such as dawn or sunset—or when the prompt demands highly dramatic atmospheric effects. These conditions require a delicate interplay of light, color gradients, and reflections, which can be difficult for generative models to reproduce in a way that remains both visually coherent and stylistically faithful.

\begin{figure}[h]
    \centering
    {\footnotesize
    \begin{tabular}{ccc}
          \includegraphics[height=0.29\linewidth]{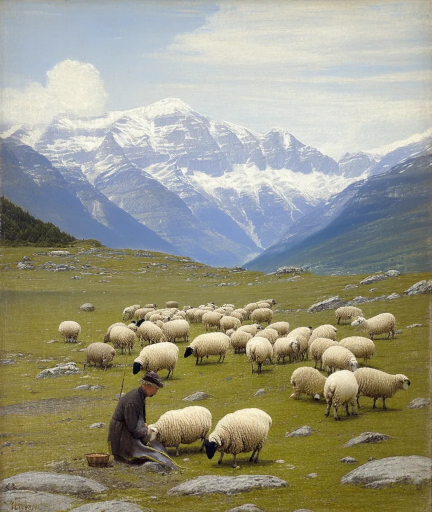} &  \includegraphics[height=0.29\linewidth]{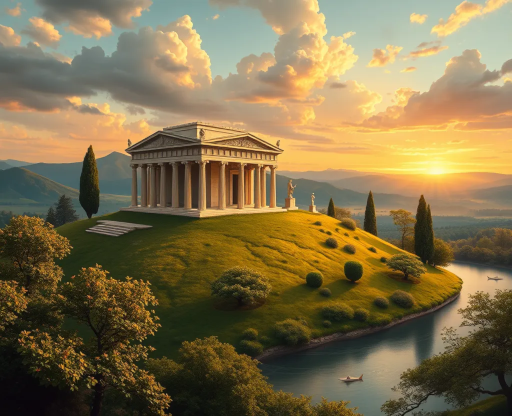} & 
          \includegraphics[height=0.29\linewidth]{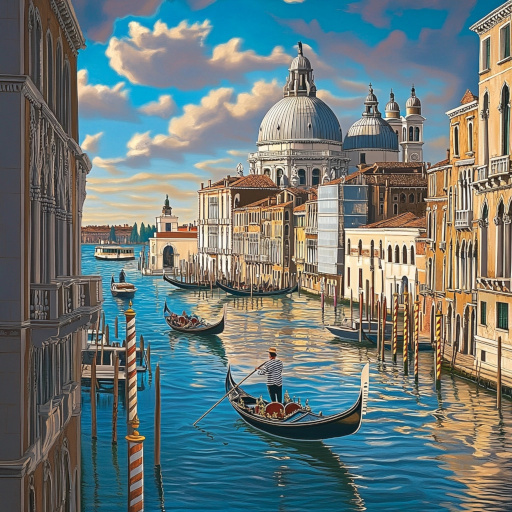}\\  
          (a) Stable-Diffusion-3.5 & (b) Flux 1.1 -Schnell & (c) Midjourney
    \end{tabular}
    }
    \caption{Hyperrealism examples. The required styles were: (a) impressionis,, (b) baroque, (c) rococo vedutism }
    \label{fig:hyperrealism_cluouds_water}
\end{figure}

Successfully depicting clouds and water often requires a nuanced understanding of texture, movement, and atmospheric perspective. While most models have made significant progress in generating these elements with a high degree of realism, they often struggle when tasked with deviating from photorealism in favor of a specific artistic technique. Instead of adapting to the  rushed brushwork of Impressionism or the soft, fairy-like contrast of Baroque landscapes, models frequently default to overly detailed or artificially blended textures, resulting in images that feel technically proficient but stylistically inconsistent with historical painting traditions.
Some examples are given in Figure~\ref{fig:hyperrealism_cluouds_water}.

The tension between hyperrealism and stylistic accuracy becomes particularly evident in the case of naïve art. While hyperrealism is defined by extreme attention to detail, the naïve style is intentionally simplified, often characterized by flat perspectives, bold colors, and a disregard for proportional accuracy. Figures and objects may appear distorted or childlike, resembling works created without adherence to formal artistic training.

\begin{figure}[h]
    \centering
    {\footnotesize
    \begin{tabular}{ccc}
        \includegraphics[width=0.3\linewidth]{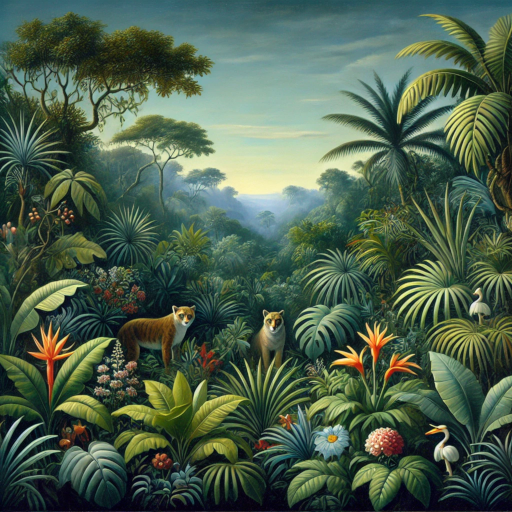} 
         & \includegraphics[width=0.3\linewidth]{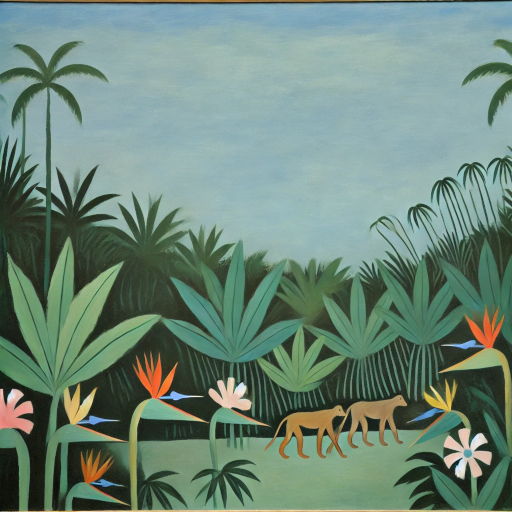} 
         &  \includegraphics[width=0.3\linewidth]{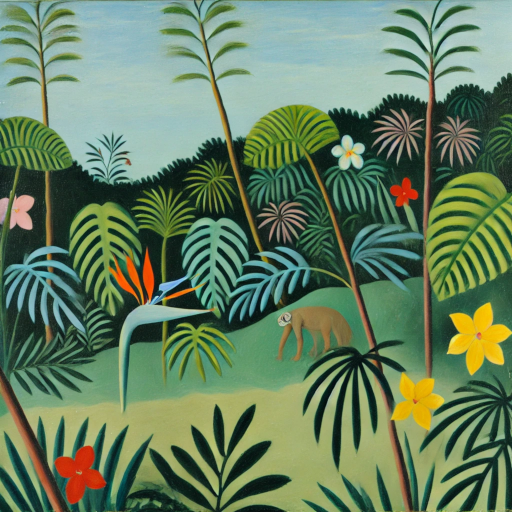}\\
         (a) & (b) & (c)
    \end{tabular}
    }
    \caption{(a) A na\"if style sample generated by Dall$\cdot$E, with an excessive sense of prospective. (b-c) Prompting 
    the model to reduce the fading effect at the horizon and produce a more flat perspective results in an oversimplification of the produced artifact.}
    \label{fig:naive_fading}
\end{figure}

An example of this contrast is shown in Figure~\ref{fig:naive_fading}(a), where the model introduces a strong fading effect at the horizon. While this technique enhances depth and improves realism, it is fundamentally at odds with the flat rendering style typical of naïve painting. When prompted to generate a flatter sky with reduced fading and perspective effects, the model tends to overcompensate, resulting in an oversimplified artifact that still fails to fully capture the intended aesthetic, as illustrated in Figures~\ref{fig:naive_fading}(b-c).

%

\subsection{Anachronisms}
\label{sec:anachronisms}
\begin{figure}[h]
    \centering
    {\footnotesize
    \begin{tabular}{ccc}
          \includegraphics[width=0.27\linewidth]{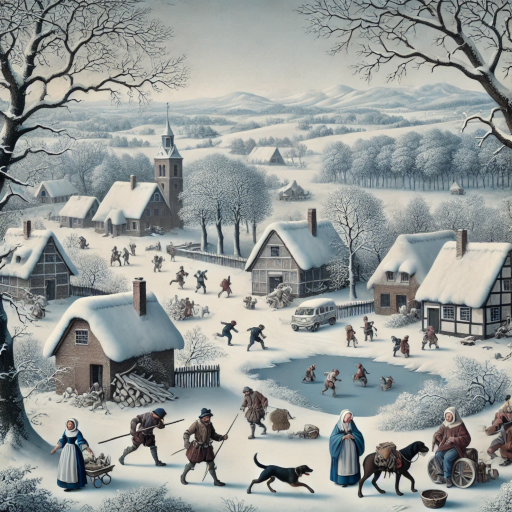} &
          \includegraphics[width=0.27\linewidth]{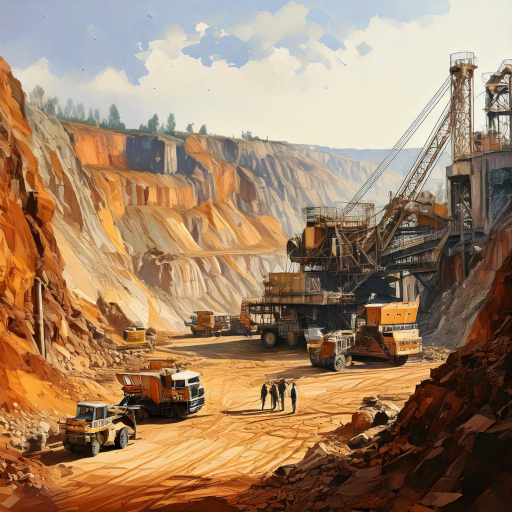} &
          \includegraphics[width=0.27\linewidth]{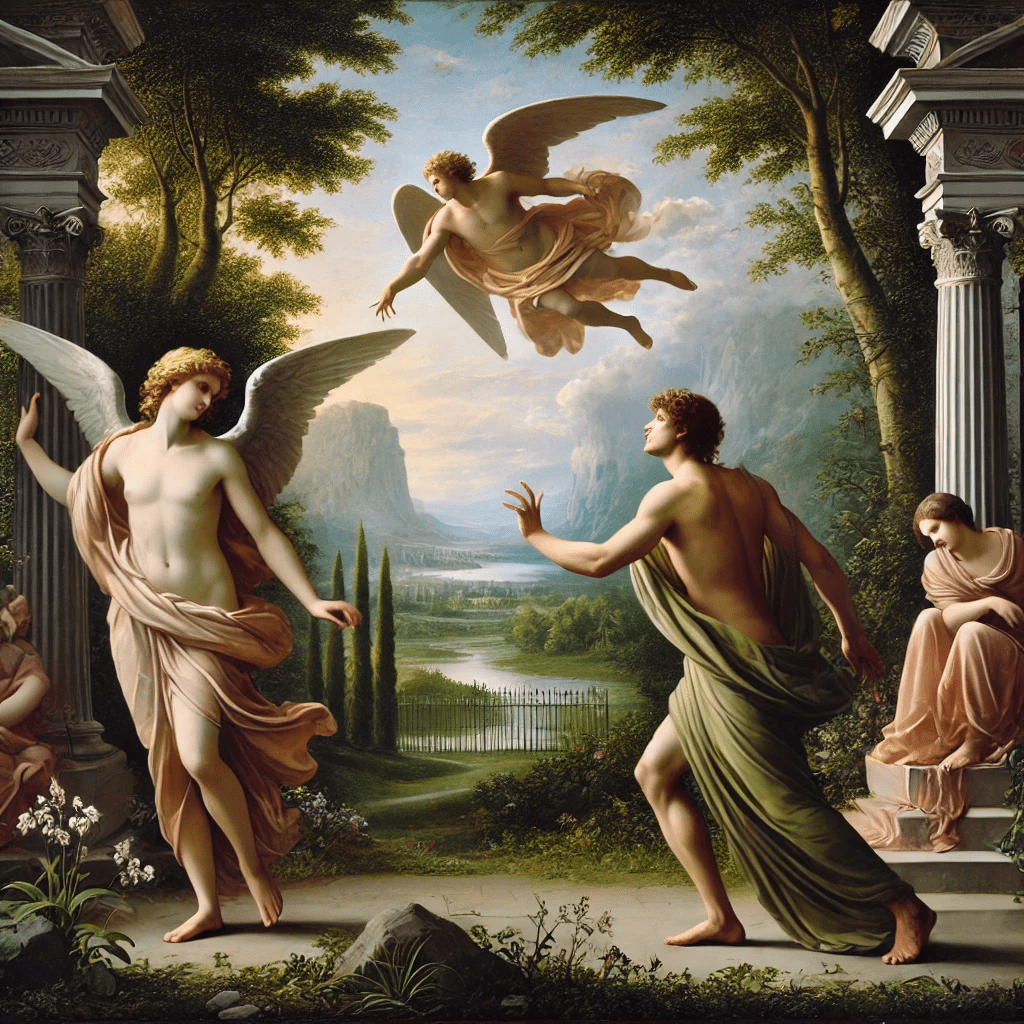} \\
          (a) \parbox[t]{3.2cm}{Dall$\cdot$E 3: Winter landscape in ``Bruegel's style''.}& (b) \parbox[t]{3.2cm}{Firefly: Realism Art of the
          Industrial Revolution period.} &
          (c) \parbox[t]{3.2cm}{Dall$\cdot$E 3: Expulsion from Eden in ``Raffaello's style''.}
    \end{tabular}
    }
    \caption{Caption}
    \label{fig:anachronisms}
\end{figure}

Not infrequently, models may add anachronistic elements in the painting,
completely disrupting the historical setting, and often creating unintentionally humorous effects. A typical example is
the van in the middle of the scene of Figure~\ref{fig:anachronisms}.a, inspired to the style of Pieter Bruegel the Elder.

The artwork of Figure~\ref{fig:anachronisms}~(b) was supposed to be a watercolor inspired by the Realism Art of the Industrial Revolution period, depicting an open-cut mine with industrial activity. Almost all models filled the scene with modern crane-like machines and trucks. 

Another example is the iron fence in Figure~\ref{fig:anachronisms}~(c). The work is intended to represent the Expulsion from the Garden of Eden, supposedly mimicking the style of Raphael. The composition is rather confused and the style is essentially neoclassical; however, the iron fence enclosing the ``garden" is particularly jarring. Wrought iron fences of this type became common in Europe only in the 19th century, primarily for gates or balconies, so its presence in a painting that is meant to belong to the Renaissance period is entirely unjustified.

\begin{figure}[h]
    \centering
    {\footnotesize
    \begin{tabular}{cc}

          \includegraphics[width=0.3\linewidth]{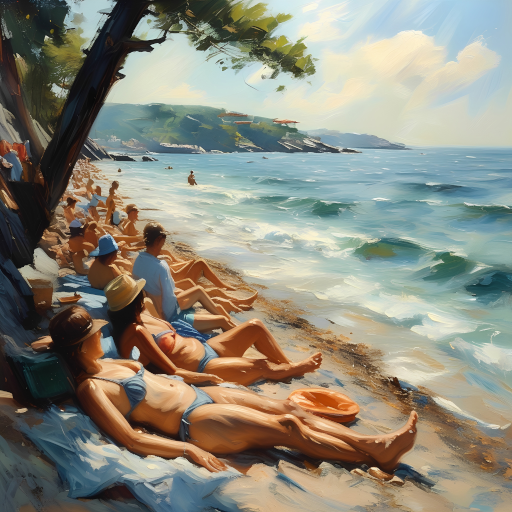}\hspace{.5cm} &  \hspace{.5cm}\includegraphics[width=0.3\linewidth]{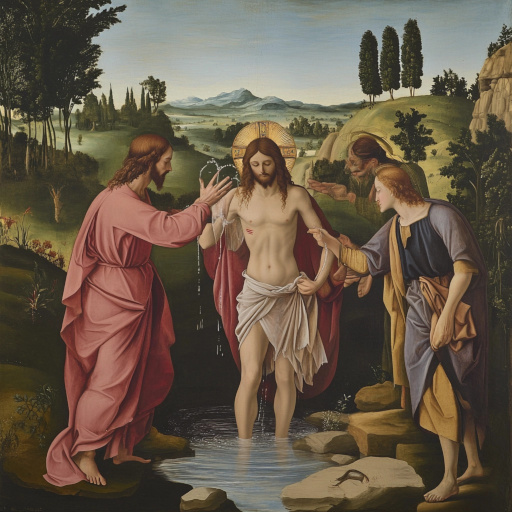} \\
          (a) Kolors: impressionist seaside & 
          (b) Midjourney: Baptism of Christ
    \end{tabular}
    }
    \caption{Caption}
    \label{fig:anachronisms2}
\end{figure}

While the clothing is generally accurate for the specified period and style, there are still some noticeable mistakes. For example, in 
Figure~\ref{fig:hyperrealism_faces}(d), the girl washing garments, ostensibly from the 19th century, is wearing sneakers — a clear anachronism. Similarly, the group of people lounging by the seaside in Figure~\ref{fig:anachronisms2}, intended to reflect the Impressionist movement, includes women wearing bikinis — an anachronism that is clearly out of place.

The most intriguing and complex example of anachronism is Midjourney's depiction of the Baptism of Christ, shown in 
Figure~\ref{fig:anachronisms2}. The issue lies in the wound on Jesus' side, traditionally associated with the crucifixion.
This example is fascinating as it highlights a fundamental challenge in generative models: the ability to semantically interpret and contextually place visual elements. The inclusion of the wound suggests the model has conflated distinct aspects of Christian iconography, likely due to overlapping representations of Christ within its training data drawn from various narratives.

This mistake underscores the difficulty of ensuring historical and theological accuracy in generative outputs, particularly when representing complex religious or cultural symbols. It reflects a lack of nuanced understanding, with the model treating all depictions of Jesus as interchangeable rather than context-specific. Addressing such issues would require either more carefully curated training data or the development of advanced mechanisms for context-aware generation. This example serves as a compelling case study in the importance of aligning generative outputs with both stylistic fidelity and semantic coherence.

\section{Conclusions}
\label{sec:conclusions}
In this work, we have explored the capabilities and limitations of modern generative models in replicating historical artistic styles. Our analysis is structured around two main contributions: (1) the creation of a large, supervised dataset of AI-generated artworks— the AI-Pastiche dataset— and (2) a comprehensive evaluation of generative models through user surveys assessing perceptual authenticity and prompt adherence.

The AI-Pastiche dataset is a richly annotated collection of AI-generated images, categorized by model, style, period, and subject matter. It serves as a valuable resource for analyzing the strengths and weaknesses of different generative approaches, holding potential for a wide range of applications and providing a benchmark for future research on AI-driven artistic replication.

Using the AI-Pastiche dataset, we conducted a systematic evaluation of generative models based on extensive user surveys. We separately assessed perceptual authenticity—how convincingly an artwork mimics human-created paintings—and prompt adherence—how faithfully the output aligns with the given instructions. The results reveal a key trade-off: some models prioritize aesthetic quality over strict adherence to the prompt, while others sacrifice visual refinement for greater accuracy. This discrepancy underscores the challenges in balancing creative flexibility and control in generative image synthesis.

Our study highlights both the progress and ongoing challenges in generative AI for artistic style replication. While models can produce visually compelling outputs, a major obstacle remains their tendency toward hyperrealism. In attempting to reproduce historical styles, these models focus on surface-level details, such as textures and brushwork, yet fail to capture the deeper artistic principles that define each period. Artistic style is more than a sum of textures—it involves composition, narrative intent, spatial relationships, and cultural context. Given the limited availability of training data for many historical styles, achieving a truly contextually accurate AI-generated artwork remains a difficult task.

Another fundamental limitation is the rigid inference time of generative models. Unlike human artists, who naturally allocate more effort to complex compositions, these models operate under fixed computational budgets, leading to missed opportunities for adaptive refinement. Future improvements may involve confidence-based step adjustments, allowing the model to extend or shorten the generation process depending on the complexity of the scene. More advanced conditioning mechanisms could also enable models to better integrate structural coherence and artistic intent rather than simply mimicking surface features.

Ultimately, our findings point to the next frontier in generative AI for art: moving beyond simple visual reproduction toward models that can understand and interpret artistic traditions in a more holistic and historically grounded way. While significant challenges remain, improvements in training strategies, dataset curation, and adaptive inference methods could help bridge the gap between style imitation and true artistic coherence, bringing generative AI closer to meaningful contributions in digital artistry.

\section*{Declarations}

\textbf{Funding.} Research partially supported by the Future AI Research (FAIR) project of the National Recovery and Resilience Plan (NRRP), Mission 4 Component 2 Investment 1.3 funded from the European Union - NextGenerationEU.\smallskip

\noindent
\textbf{Conflict of Interest.} On behalf of all authors, the corresponding author states that there is no conflict of interest.



\begin{thebibliography}{47}
\ifx \bisbn   \undefined \def \bisbn  #1{ISBN #1}\fi
\ifx \binits  \undefined \def \binits#1{#1}\fi
\ifx \bauthor  \undefined \def \bauthor#1{#1}\fi
\ifx \batitle  \undefined \def \batitle#1{#1}\fi
\ifx \bjtitle  \undefined \def \bjtitle#1{#1}\fi
\ifx \bvolume  \undefined \def \bvolume#1{\textbf{#1}}\fi
\ifx \byear  \undefined \def \byear#1{#1}\fi
\ifx \bissue  \undefined \def \bissue#1{#1}\fi
\ifx \bfpage  \undefined \def \bfpage#1{#1}\fi
\ifx \blpage  \undefined \def \blpage #1{#1}\fi
\ifx \burl  \undefined \def \burl#1{\textsf{#1}}\fi
\ifx \doiurl  \undefined \def \doiurl#1{\url{https://doi.org/#1}}\fi
\ifx \betal  \undefined \def \betal{\textit{et al.}}\fi
\ifx \binstitute  \undefined \def \binstitute#1{#1}\fi
\ifx \binstitutionaled  \undefined \def \binstitutionaled#1{#1}\fi
\ifx \bctitle  \undefined \def \bctitle#1{#1}\fi
\ifx \beditor  \undefined \def \beditor#1{#1}\fi
\ifx \bpublisher  \undefined \def \bpublisher#1{#1}\fi
\ifx \bbtitle  \undefined \def \bbtitle#1{#1}\fi
\ifx \bedition  \undefined \def \bedition#1{#1}\fi
\ifx \bseriesno  \undefined \def \bseriesno#1{#1}\fi
\ifx \blocation  \undefined \def \blocation#1{#1}\fi
\ifx \bsertitle  \undefined \def \bsertitle#1{#1}\fi
\ifx \bsnm \undefined \def \bsnm#1{#1}\fi
\ifx \bsuffix \undefined \def \bsuffix#1{#1}\fi
\ifx \bparticle \undefined \def \bparticle#1{#1}\fi
\ifx \barticle \undefined \def \barticle#1{#1}\fi
\bibcommenthead
\ifx \bconfdate \undefined \def \bconfdate #1{#1}\fi
\ifx \botherref \undefined \def \botherref #1{#1}\fi
\ifx \url \undefined \def \url#1{\textsf{#1}}\fi
\ifx \bchapter \undefined \def \bchapter#1{#1}\fi
\ifx \bbook \undefined \def \bbook#1{#1}\fi
\ifx \bcomment \undefined \def \bcomment#1{#1}\fi
\ifx \oauthor \undefined \def \oauthor#1{#1}\fi
\ifx \citeauthoryear \undefined \def \citeauthoryear#1{#1}\fi
\ifx \endbibitem  \undefined \def \endbibitem {}\fi
\ifx \bconflocation  \undefined \def \bconflocation#1{#1}\fi
\ifx \arxivurl  \undefined \def \arxivurl#1{\textsf{#1}}\fi
\csname PreBibitemsHook\endcsname

\bibitem{StyleGAN2}
\begin{bchapter}
\bauthor{\bsnm{Karras}, \binits{T.}},
\bauthor{\bsnm{Laine}, \binits{S.}},
\bauthor{\bsnm{Aittala}, \binits{M.}},
\bauthor{\bsnm{Hellsten}, \binits{J.}},
\bauthor{\bsnm{Lehtinen}, \binits{J.}},
\bauthor{\bsnm{Aila}, \binits{T.}}:
\bctitle{Analyzing and improving the image quality of stylegan}.
In: \bbtitle{2020 {IEEE/CVF} Conference on Computer Vision and Pattern Recognition, {CVPR} 2020, Seattle, WA, USA, June 13-19, 2020},
pp. \bfpage{8107}--\blpage{8116}.
\bpublisher{Computer Vision Foundation / {IEEE}},
\blocation{\hspace{0pt}}
(\byear{2020}).
\doiurl{10.1109/CVPR42600.2020.00813}
\end{bchapter}
\endbibitem

\bibitem{StyleGAN-T}
\begin{bchapter}
\bauthor{\bsnm{Sauer}, \binits{A.}},
\bauthor{\bsnm{Karras}, \binits{T.}},
\bauthor{\bsnm{Laine}, \binits{S.}},
\bauthor{\bsnm{Geiger}, \binits{A.}},
\bauthor{\bsnm{Aila}, \binits{T.}}:
\bctitle{Stylegan-t: Unlocking the power of gans for fast large-scale text-to-image synthesis}.
In: \beditor{\bsnm{Krause}, \binits{A.}},
\beditor{\bsnm{Brunskill}, \binits{E.}},
\beditor{\bsnm{Cho}, \binits{K.}},
\beditor{\bsnm{Engelhardt}, \binits{B.}},
\beditor{\bsnm{Sabato}, \binits{S.}},
\beditor{\bsnm{Scarlett}, \binits{J.}} (eds.)
\bbtitle{International Conference on Machine Learning, {ICML} 2023, 23-29 July 2023, Honolulu, Hawaii, {USA}}.
\bsertitle{Proceedings of Machine Learning Research},
vol. \bseriesno{202},
pp. \bfpage{30105}--\blpage{30118}.
\bpublisher{{PMLR}},
\blocation{\hspace{0pt}}
(\byear{2023}).
\bcomment{Accessed: 2025-02-13}.
\burl{https://proceedings.mlr.press/v202/sauer23a.html}
\end{bchapter}
\endbibitem

\bibitem{DDPM}
\begin{bchapter}
\bauthor{\bsnm{Ho}, \binits{J.}},
\bauthor{\bsnm{Jain}, \binits{A.}},
\bauthor{\bsnm{Abbeel}, \binits{P.}}:
\bctitle{Denoising diffusion probabilistic models}.
In: \beditor{\bsnm{Larochelle}, \binits{H.}},
\beditor{\bsnm{Ranzato}, \binits{M.}},
\beditor{\bsnm{Hadsell}, \binits{R.}},
\beditor{\bsnm{Balcan}, \binits{M.}},
\beditor{\bsnm{Lin}, \binits{H.}} (eds.)
\bbtitle{Advances in Neural Information Processing Systems 33: Annual Conference on Neural Information Processing Systems 2020, NeurIPS 2020, December 6-12, 2020, Virtual}
(\byear{2020}).
\bcomment{Accessed: 2025-02-13}.
\burl{https://proceedings.neurips.cc/paper/2020/hash/4c5bcfec8584af0d967f1ab10179ca4b-Abstract.html}
\end{bchapter}
\endbibitem

\bibitem{DDIM}
\begin{bchapter}
\bauthor{\bsnm{Song}, \binits{J.}},
\bauthor{\bsnm{Meng}, \binits{C.}},
\bauthor{\bsnm{Ermon}, \binits{S.}}:
\bctitle{Denoising diffusion implicit models}.
In: \bbtitle{9th International Conference on Learning Representations, {ICLR} 2021, Virtual Event, Austria, May 3-7, 2021}.
\bpublisher{OpenReview.net},
\blocation{\hspace{0pt}}
(\byear{2021}).
\bcomment{Accessed: 2025-02-13}.
\burl{https://openreview.net/forum?id=St1giarCHLP}
\end{bchapter}
\endbibitem

\bibitem{stable-diffusion}
\begin{bchapter}
\bauthor{\bsnm{Rombach}, \binits{R.}},
\bauthor{\bsnm{Blattmann}, \binits{A.}},
\bauthor{\bsnm{Lorenz}, \binits{D.}},
\bauthor{\bsnm{Esser}, \binits{P.}},
\bauthor{\bsnm{Ommer}, \binits{B.}}:
\bctitle{High-resolution image synthesis with latent diffusion models}.
In: \bbtitle{Proceedings of the IEEE/CVF Conference on Computer Vision and Pattern Recognition},
pp. \bfpage{10684}--\blpage{10695}
(\byear{2022})
\end{bchapter}
\endbibitem

\bibitem{SGMD}
\begin{barticle}
\bauthor{\bsnm{Xu}, \binits{Y.}},
\bauthor{\bsnm{Xu}, \binits{X.}},
\bauthor{\bsnm{Gao}, \binits{H.}},
\bauthor{\bsnm{Xiao}, \binits{F.}}:
\batitle{{SGDM:} an adaptive style-guided diffusion model for personalized text to image generation}.
\bjtitle{{IEEE} Trans. Multim.}
\bvolume{26},
\bfpage{9804}--\blpage{9813}
(\byear{2024}).
\doiurl{10.1109/TMM.2024.3399075}
\end{barticle}
\endbibitem

\bibitem{ArtBank}
\begin{bchapter}
\bauthor{\bsnm{Zhang}, \binits{Z.}},
\bauthor{\bsnm{Zhang}, \binits{Q.}},
\bauthor{\bsnm{Xing}, \binits{W.}},
\bauthor{\bsnm{Li}, \binits{G.}},
\bauthor{\bsnm{Zhao}, \binits{L.}},
\bauthor{\bsnm{Sun}, \binits{J.}},
\bauthor{\bsnm{Lan}, \binits{Z.}},
\bauthor{\bsnm{Luan}, \binits{J.}},
\bauthor{\bsnm{Huang}, \binits{Y.}},
\bauthor{\bsnm{Lin}, \binits{H.}}:
\bctitle{Artbank: Artistic style transfer with pre-trained diffusion model and implicit style prompt bank}.
In: \bbtitle{Thirty-Eighth {AAAI} Conference on Artificial Intelligence, {AAAI} 2024, Thirty-Sixth Conference on Innovative Applications of Artificial Intelligence, {IAAI} 2024, Fourteenth Symposium on Educational Advances in Artificial Intelligence, {EAAI} 2014, February 20-27, 2024, Vancouver, Canada},
pp. \bfpage{7396}--\blpage{7404}
(\byear{2024}).
\doiurl{10.1609/AAAI.V38I7.28570}
\end{bchapter}
\endbibitem

\bibitem{Dalle2}
\begin{botherref}
\oauthor{\bsnm{Ramesh}, \binits{A.}},
\oauthor{\bsnm{Dhariwal}, \binits{P.}},
\oauthor{\bsnm{Nichol}, \binits{A.}},
\oauthor{\bsnm{Chu}, \binits{C.}},
\oauthor{\bsnm{Chen}, \binits{M.}}:
Hierarchical text-conditional image generation with clip latents.
arXiv e-prints,
2204
(2022)
\end{botherref}
\endbibitem

\bibitem{Imagen}
\begin{bchapter}
\bauthor{\bsnm{Saharia}, \binits{C.}},
\bauthor{\bsnm{Chan}, \binits{W.}},
\bauthor{\bsnm{Saxena}, \binits{S.}},
\bauthor{\bsnm{Li}, \binits{L.}},
\bauthor{\bsnm{Whang}, \binits{J.}},
\bauthor{\bsnm{Denton}, \binits{E.L.}},
\bauthor{\bsnm{Ghasemipour}, \binits{S.K.S.}},
\bauthor{\bsnm{Lopes}, \binits{R.G.}},
\bauthor{\bsnm{Ayan}, \binits{B.K.}},
\bauthor{\bsnm{Salimans}, \binits{T.}},
\bauthor{\bsnm{Ho}, \binits{J.}},
\bauthor{\bsnm{Fleet}, \binits{D.J.}},
\bauthor{\bsnm{Norouzi}, \binits{M.}}:
\bctitle{Photorealistic text-to-image diffusion models with deep language understanding}.
In: \bbtitle{NeurIPS}
(\byear{2022}).
\bcomment{Accessed: 2025-02-13}.
\burl{http://papers.nips.cc/paper\_files/paper/2022/hash/ec795aeadae0b7d230fa35cbaf04c041-Abstract-Conference.html}
\end{bchapter}
\endbibitem

\bibitem{FeaST}
\begin{barticle}
\bauthor{\bsnm{Png}, \binits{W.H.}},
\bauthor{\bsnm{Aun}, \binits{Y.}},
\bauthor{\bsnm{Gan}, \binits{M.}}:
\batitle{Feast: Feature-guided style transfer for high-fidelity art synthesis}.
\bjtitle{Comput. Graph.}
\bvolume{122},
\bfpage{103975}
(\byear{2024}).
\doiurl{10.1016/J.CAG.2024.103975}
\end{barticle}
\endbibitem

\bibitem{Style_injection}
\begin{bchapter}
\bauthor{\bsnm{Chung}, \binits{J.}},
\bauthor{\bsnm{Hyun}, \binits{S.}},
\bauthor{\bsnm{Heo}, \binits{J.}}:
\bctitle{Style injection in diffusion: {A} training-free approach for adapting large-scale diffusion models for style transfer}.
In: \bbtitle{{IEEE/CVF} Conference on Computer Vision and Pattern Recognition, {CVPR} 2024, Seattle, WA, USA, June 16-22, 2024},
pp. \bfpage{8795}--\blpage{8805}
(\byear{2024}).
\doiurl{10.1109/CVPR52733.2024.00840}
\end{bchapter}
\endbibitem

\bibitem{Step-aware}
\begin{bchapter}
\bauthor{\bsnm{Zhang}, \binits{Z.}},
\bauthor{\bsnm{Zhang}, \binits{Q.}},
\bauthor{\bsnm{Lin}, \binits{H.}},
\bauthor{\bsnm{Xing}, \binits{W.}},
\bauthor{\bsnm{Mo}, \binits{J.}},
\bauthor{\bsnm{Huang}, \binits{S.}},
\bauthor{\bsnm{Xie}, \binits{J.}},
\bauthor{\bsnm{Li}, \binits{G.}},
\bauthor{\bsnm{Luan}, \binits{J.}},
\bauthor{\bsnm{Zhao}, \binits{L.}},
\bauthor{\bsnm{Zhang}, \binits{D.}},
\bauthor{\bsnm{Chen}, \binits{L.}}:
\bctitle{Towards highly realistic artistic style transfer via stable diffusion with step-aware and layer-aware prompt}.
In: \bbtitle{Proceedings of the Thirty-Third International Joint Conference on Artificial Intelligence, {IJCAI} 2024, Jeju, South Korea, August 3-9, 2024},
pp. \bfpage{7814}--\blpage{7822}
(\byear{2024}).
\bcomment{Accessed: 2025-02-13}.
\burl{https://www.ijcai.org/proceedings/2024/865}
\end{bchapter}
\endbibitem

\bibitem{designing-barros}
\begin{barticle}
\bauthor{\bsnm{Barros}, \binits{M.}},
\bauthor{\bsnm{Ai}, \binits{Q.}}:
\batitle{Designing with words: exploring the integration of text-to-image models in industrial design}.
\bjtitle{Digit. Creativity}
\bvolume{35}(\bissue{4}),
\bfpage{378}--\blpage{391}
(\byear{2024}).
\doiurl{10.1080/14626268.2024.2411223}
\end{barticle}
\endbibitem

\bibitem{GameDesign}
\begin{bchapter}
\bauthor{\bsnm{Zhou}, \binits{H.}},
\bauthor{\bsnm{Zhu}, \binits{J.}},
\bauthor{\bsnm{Mateas}, \binits{M.}},
\bauthor{\bsnm{Wardrip{-}Fruin}, \binits{N.}}:
\bctitle{The eyes, the hands and the brain: What can text-to-image models offer for game design and visual creativity?}
In: \beditor{\bsnm{Smith}, \binits{G.}},
\beditor{\bsnm{Whitehead}, \binits{J.}},
\beditor{\bsnm{Samuel}, \binits{B.}},
\beditor{\bsnm{Spiel}, \binits{K.}},
\beditor{\bparticle{van} \bsnm{Rozen}, \binits{R.}} (eds.)
\bbtitle{Proceedings of the 19th International Conference on the Foundations of Digital Games, {FDG} 2024, Worcester, MA, USA, May 21-24, 2024},
p. \bfpage{23}.
\bpublisher{{ACM}},
\blocation{\hspace{0pt}}
(\byear{2024}).
\bcomment{Accessed: 2025-02-13}.
\burl{https://doi.org/10.1145/3649921.3650001}
\end{bchapter}
\endbibitem

\bibitem{inspiring-Haase}
\begin{bchapter}
\bauthor{\bsnm{Haase}, \binits{J.}},
\bauthor{\bsnm{Djurica}, \binits{D.}},
\bauthor{\bsnm{Mendling}, \binits{J.}}:
\bctitle{The art of inspiring creativity: Exploring the unique impact of ai-generated images}.
In: \beditor{\bsnm{Pavlou}, \binits{P.A.}},
\beditor{\bsnm{Midha}, \binits{V.}},
\beditor{\bsnm{Animesh}, \binits{A.}},
\beditor{\bsnm{Carte}, \binits{T.A.}},
\beditor{\bsnm{Graeml}, \binits{A.R.}},
\beditor{\bsnm{Mitchell}, \binits{A.}} (eds.)
\bbtitle{29th Americas Conference on Information Systems, {AMCIS} 2023, Panama City, Panama, August 10-12, 2023}.
\bpublisher{Association for Information Systems},
\blocation{\hspace{0pt}}
(\byear{2023}).
\bcomment{Accessed: 2025-02-13}.
\burl{https://aisel.aisnet.org/amcis2023/sig\_aiaa/sig\_aiaa/10}
\end{bchapter}
\endbibitem

\bibitem{craft-edu}
\begin{barticle}
\bauthor{\bsnm{Vartiainen}, \binits{H.}},
\bauthor{\bsnm{Tedre}, \binits{M.}}:
\batitle{Using artificial intelligence in craft education: crafting with text-to-image generative models}.
\bjtitle{Digit. Creativity}
\bvolume{34}(\bissue{1}),
\bfpage{1}--\blpage{21}
(\byear{2023}).
\bcomment{Accessed: 2025-02-13}
\end{barticle}
\endbibitem

\bibitem{emergent}
\begin{botherref}
\oauthor{\bsnm{Wei}, \binits{J.}},
\oauthor{\bsnm{Tay}, \binits{Y.}},
\oauthor{\bsnm{Bommasani}, \binits{R.}},
\oauthor{\bsnm{Raffel}, \binits{C.}},
\oauthor{\bsnm{Zoph}, \binits{B.}},
\oauthor{\bsnm{Borgeaud}, \binits{S.}},
\oauthor{\bsnm{Yogatama}, \binits{D.}},
\oauthor{\bsnm{Bosma}, \binits{M.}},
\oauthor{\bsnm{Zhou}, \binits{D.}},
\oauthor{\bsnm{Metzler}, \binits{D.}},
\oauthor{\bsnm{Chi}, \binits{E.H.}},
\oauthor{\bsnm{Hashimoto}, \binits{T.}},
\oauthor{\bsnm{Vinyals}, \binits{O.}},
\oauthor{\bsnm{Liang}, \binits{P.}},
\oauthor{\bsnm{Dean}, \binits{J.}},
\oauthor{\bsnm{Fedus}, \binits{W.}}:
Emergent abilities of large language models.
Trans. Mach. Learn. Res.
\textbf{2022}
(2022).
Accessed: 2025-02-13
\end{botherref}
\endbibitem

\bibitem{emergent_mirage}
\begin{bchapter}
\bauthor{\bsnm{Schaeffer}, \binits{R.}},
\bauthor{\bsnm{Miranda}, \binits{B.}},
\bauthor{\bsnm{Koyejo}, \binits{S.}}:
\bctitle{Are emergent abilities of large language models a mirage?}
In: \bbtitle{Advances in Neural Information Processing Systems 36: Annual Conference on Neural Information Processing Systems 2023, NeurIPS 2023, New Orleans, LA, USA, December 10 - 16, 2023}
(\byear{2023}).
\bcomment{Accessed: 2025-02-13}.
\burl{http://papers.nips.cc/paper\_files/paper/2023/hash/adc98a266f45005c403b8311ca7e8bd7-Abstract-Conference.html}
\end{bchapter}
\endbibitem

\bibitem{loss_perspective}
\begin{botherref}
\oauthor{\bsnm{Du}, \binits{Z.}},
\oauthor{\bsnm{Zeng}, \binits{A.}},
\oauthor{\bsnm{Dong}, \binits{Y.}},
\oauthor{\bsnm{Tang}, \binits{J.}}:
Understanding emergent abilities of language models from the loss perspective.
CoRR
\textbf{abs/2403.15796}
(2024).
\doiurl{10.48550/ARXIV.2403.15796}
\end{botherref}
\endbibitem

\bibitem{Angelo_TOM}
\begin{bchapter}
\bauthor{\bsnm{Huang}, \binits{X.A.}},
\bauthor{\bsnm{Malfa}, \binits{E.L.}},
\bauthor{\bsnm{Marro}, \binits{S.}},
\bauthor{\bsnm{Asperti}, \binits{A.}},
\bauthor{\bsnm{Cohn}, \binits{A.G.}},
\bauthor{\bsnm{Wooldridge}, \binits{M.J.}}:
\bctitle{A notion of complexity for theory of mind via discrete world models}.
In: \beditor{\bsnm{Al{-}Onaizan}, \binits{Y.}},
\beditor{\bsnm{Bansal}, \binits{M.}},
\beditor{\bsnm{Chen}, \binits{Y.}} (eds.)
\bbtitle{Findings of the Association for Computational Linguistics: {EMNLP} 2024, Miami, Florida, USA, November 12-16, 2024},
pp. \bfpage{2964}--\blpage{2983}.
\bpublisher{{Association for Computational Linguistics}},
\blocation{\hspace{0pt}}
(\byear{2024}).
\burl{https://aclanthology.org/2024.findings-emnlp.167}
\end{bchapter}
\endbibitem

\bibitem{Gatys_neural_representation}
\begin{botherref}
\oauthor{\bsnm{Gatys}, \binits{L.A.}},
\oauthor{\bsnm{Ecker}, \binits{A.S.}},
\oauthor{\bsnm{Bethge}, \binits{M.}}:
A neural algorithm of artistic style.
CoRR
\textbf{abs/1508.06576}
(2015)
{\href{https://arxiv.org/abs/1508.06576}{{1508.06576}}}.
Accessed: 2025-02-13
\end{botherref}
\endbibitem

\bibitem{Huang_style_transfer_normalization}
\begin{bchapter}
\bauthor{\bsnm{Huang}, \binits{X.}},
\bauthor{\bsnm{Belongie}, \binits{S.J.}}:
\bctitle{Arbitrary style transfer in real-time with adaptive instance normalization}.
In: \bbtitle{{IEEE} International Conference on Computer Vision, {ICCV} 2017, Venice, Italy, October 22-29, 2017},
pp. \bfpage{1510}--\blpage{1519}.
\bpublisher{{IEEE} Computer Society},
\blocation{\hspace{0pt}}
(\byear{2017}).
\doiurl{10.1109/ICCV.2017.167}
\end{bchapter}
\endbibitem

\bibitem{cifake}
\begin{barticle}
\bauthor{\bsnm{Bird}, \binits{J.J.}},
\bauthor{\bsnm{Lotfi}, \binits{A.}}:
\batitle{Cifake: Image classification and explainable identification of ai-generated synthetic images}.
\bjtitle{CoRR}
(\byear{2023})
{\href{https://arxiv.org/abs/2303.14126}{{2303.14126}}}.
\doiurl{arXiv:2303.14126}
\end{barticle}
\endbibitem

\bibitem{genimage}
\begin{barticle}
\bauthor{\bsnm{Zhu}, \binits{M.}},
\bauthor{\bsnm{Chen}, \binits{H.}},
\bauthor{\bsnm{Yan}, \binits{Q.}},
\bauthor{\bsnm{Huang}, \binits{X.}},
\bauthor{\bsnm{Lin}, \binits{G.}},
\bauthor{\bsnm{Li}, \binits{W.}},
\bauthor{\bsnm{Tuv}, \binits{Z.}},
\bauthor{\bsnm{Hu}, \binits{H.}},
\bauthor{\bsnm{Hu}, \binits{J.}},
\bauthor{\bsnm{Wang}, \binits{Y.}}:
\batitle{Genimage: A million-scale benchmark for detecting ai-generated image}.
\bjtitle{CoRR}
(\byear{2023}).
\doiurl{arXiv:2306.0857}
\end{barticle}
\endbibitem

\bibitem{AIart}
\begin{barticle}
\bauthor{\bsnm{Li}, \binits{Y.}},
\bauthor{\bsnm{Liu}, \binits{Z.}},
\bauthor{\bsnm{Zhao}, \binits{J.}},
\bauthor{\bsnm{Ren}, \binits{L.}},
\bauthor{\bsnm{Li}, \binits{F.}},
\bauthor{\bsnm{Luo}, \binits{J.}},
\bauthor{\bsnm{Luo}, \binits{B.}}:
\batitle{The adversarial ai-art: Understanding, generation, detection, and benchmarking}.
\bjtitle{CoRR}
(\byear{2024}).
\doiurl{arXiv:2404.14581}
\end{barticle}
\endbibitem

\bibitem{artifact_dataset}
\begin{bchapter}
\bauthor{\bsnm{Rahman}, \binits{M.A.}},
\bauthor{\bsnm{Paul}, \binits{B.}},
\bauthor{\bsnm{Sarker}, \binits{N.H.}},
\bauthor{\bsnm{Hakim}, \binits{Z.I.A.}},
\bauthor{\bsnm{Fattah}, \binits{S.A.}}:
\bctitle{Artifact: {A} large-scale dataset with artificial and factual images for generalizable and robust synthetic image detection}.
In: \bbtitle{{IEEE} International Conference on Image Processing, {ICIP} 2023, Kuala Lumpur, Malaysia, October 8-11, 2023},
pp. \bfpage{2200}--\blpage{2204}.
\bpublisher{{IEEE}},
\blocation{\hspace{0pt}}
(\byear{2023}).
\doiurl{10.1109/ICIP49359.2023.10222083}
\end{bchapter}
\endbibitem

\bibitem{wildfake_dataset}
\begin{botherref}
\oauthor{\bsnm{Hong}, \binits{Y.}},
\oauthor{\bsnm{Zhang}, \binits{J.}}:
Wildfake: {A} large-scale challenging dataset for ai-generated images detection.
CoRR
\textbf{abs/2402.11843}
(2024).
\doiurl{10.48550/ARXIV.2402.11843}
\end{botherref}
\endbibitem

\bibitem{twigma_dataset}
\begin{bchapter}
\bauthor{\bsnm{Chen}, \binits{Y.T.}},
\bauthor{\bsnm{Zou}, \binits{J.Y.}}:
\bctitle{{TWIGMA:} {A} dataset of ai-generated images with metadata from twitter}.
In: \beditor{\bsnm{Oh}, \binits{A.}},
\beditor{\bsnm{Naumann}, \binits{T.}},
\beditor{\bsnm{Globerson}, \binits{A.}},
\beditor{\bsnm{Saenko}, \binits{K.}},
\beditor{\bsnm{Hardt}, \binits{M.}},
\beditor{\bsnm{Levine}, \binits{S.}} (eds.)
\bbtitle{Advances in Neural Information Processing Systems 36: Annual Conference on Neural Information Processing Systems 2023, NeurIPS 2023, New Orleans, LA, USA, December 10 - 16, 2023}
(\byear{2023}).
\bcomment{Accessed: 2025-02-13}.
\burl{http://papers.nips.cc/paper\_files/paper/2023/hash/769b70d1a9a6b21af53c00d0b322c763-Abstract-Datasets\_and\_Benchmarks.html}
\end{bchapter}
\endbibitem

\bibitem{aigen}
\begin{barticle}
\bauthor{\bsnm{Yang}, \binits{Z.}},
\bauthor{\bsnm{Zhan}, \binits{F.}},
\bauthor{\bsnm{Liu}, \binits{K.}},
\bauthor{\bsnm{Xu}, \binits{M.}},
\bauthor{\bsnm{Lu}, \binits{S.}}:
\batitle{Ai-generated images as data source: The dawn of synthetic era}.
\bjtitle{CoRR}
(\byear{2023}).
\doiurl{arXiv:2310.01830}
\end{barticle}
\endbibitem

\bibitem{evaluation-metrics}
\begin{botherref}
\oauthor{\bsnm{Wang}, \binits{B.}},
\oauthor{\bsnm{Zhu}, \binits{Y.}},
\oauthor{\bsnm{Chen}, \binits{L.}},
\oauthor{\bsnm{Liu}, \binits{J.}},
\oauthor{\bsnm{Sun}, \binits{L.}},
\oauthor{\bsnm{Childs}, \binits{P.R.N.}}:
A study of the evaluation metrics for generative images containing combinational creativity.
Artif. Intell. Eng. Des. Anal. Manuf.
\textbf{37}
(2023).
Accessed: 2025-02-13
\end{botherref}
\endbibitem

\bibitem{prompts-Oppenlaendr}
\begin{bchapter}
\bauthor{\bsnm{Oppenlaender}, \binits{J.}}:
\bctitle{The creativity of text-to-image generation}.
In: \bbtitle{25th International Academic Mindtrek Conference, Academic Mindtrek 2022, Tampere, Finland, November 16-18, 2022},
pp. \bfpage{192}--\blpage{202}.
\bpublisher{{ACM}},
\blocation{\hspace{0pt}}
(\byear{2022}).
\doiurl{10.1145/3569219.3569352}
\end{bchapter}
\endbibitem

\bibitem{prompts-Sanchez23}
\begin{bchapter}
\bauthor{\bsnm{Sanchez}, \binits{T.}}:
\bctitle{Examining the text-to-image community of practice: Why and how do people prompt generative ais?}
In: \bbtitle{Creativity and Cognition, C{\&}C 2023, Virtual Event, USA, June 19-21, 2023},
pp. \bfpage{43}--\blpage{61}.
\bpublisher{{ACM}},
\blocation{\hspace{0pt}}
(\byear{2023}).
\bcomment{Accessed: 2025-02-13}.
\burl{https://doi.org/10.1145/3591196.3593051}
\end{bchapter}
\endbibitem

\bibitem{optimizing-prompts}
\begin{barticle}
\bauthor{\bsnm{Lee}, \binits{S.}},
\bauthor{\bsnm{Lee}, \binits{J.}},
\bauthor{\bsnm{Bae}, \binits{C.H.}},
\bauthor{\bsnm{Choi}, \binits{M.}},
\bauthor{\bsnm{Lee}, \binits{R.}},
\bauthor{\bsnm{Ahn}, \binits{S.}}:
\batitle{Optimizing prompts using in-context few-shot learning for text-to-image generative models}.
\bjtitle{{IEEE} Access}
\bvolume{12},
\bfpage{2660}--\blpage{2673}
(\byear{2024}).
\bcomment{Accessed: 2025-02-13}
\end{barticle}
\endbibitem

\bibitem{instruct_pix2pix_i2i}
\begin{botherref}
\oauthor{\bsnm{Brooks}, \binits{T.}},
\oauthor{\bsnm{Holynski}, \binits{A.}},
\oauthor{\bsnm{Efros}, \binits{A.A.}}:
Instructpix2pix: Learning to follow image editing instructions.
CoRR
\textbf{abs/2211.09800}
(2022).
\doiurl{10.48550/ARXIV.2211.09800}
\end{botherref}
\endbibitem

\bibitem{img2img_turbo_i2i}
\begin{botherref}
\oauthor{\bsnm{Parmar}, \binits{G.}},
\oauthor{\bsnm{Park}, \binits{T.}},
\oauthor{\bsnm{Narasimhan}, \binits{S.}},
\oauthor{\bsnm{Zhu}, \binits{J.}}:
One-step image translation with text-to-image models.
CoRR
\textbf{abs/2403.12036}
(2024).
\doiurl{10.48550/ARXIV.2403.12036}
\end{botherref}
\endbibitem

\bibitem{stabilityai2023}
\begin{botherref}
\oauthor{\bsnm{AI}, \binits{S.}}:
Stability AI releases DeepFloyd IF, a powerful text-to-image model that can smartly integrate text into images.
Accessed: 2024-11-18
(2023).
\url{https://stability.ai/news/deepfloyd-if-text-to-image-model}
\end{botherref}
\endbibitem

\bibitem{peebles2023scalablediffusionmodelstransformers}
\begin{bchapter}
\bauthor{\bsnm{Peebles}, \binits{W.}},
\bauthor{\bsnm{Xie}, \binits{S.}}:
\bctitle{Scalable diffusion models with transformers}.
In: \bbtitle{{IEEE/CVF} International Conference on Computer Vision, {ICCV} 2023, Paris, France, October 1-6, 2023},
pp. \bfpage{4172}--\blpage{4182}
(\byear{2023}).
\bcomment{Accessed: 2025-02-13}.
\burl{https://doi.org/10.1109/ICCV51070.2023.00387}
\end{bchapter}
\endbibitem

\bibitem{esser2024scalingrectifiedflowtransformers}
\begin{bchapter}
\bauthor{\bsnm{Esser}, \binits{P.}},
\bauthor{\bsnm{Kulal}, \binits{S.}},
\bauthor{\bsnm{Blattmann}, \binits{A.}},
\bauthor{\bsnm{Entezari}, \binits{R.}},
\bauthor{\bsnm{M{\"{u}}ller}, \binits{J.}},
\bauthor{\bsnm{Saini}, \binits{H.}},
\bauthor{\bsnm{Levi}, \binits{Y.}},
\bauthor{\bsnm{Lorenz}, \binits{D.}},
\bauthor{\bsnm{Sauer}, \binits{A.}},
\bauthor{\bsnm{Boesel}, \binits{F.}},
\bauthor{\bsnm{Podell}, \binits{D.}},
\bauthor{\bsnm{Dockhorn}, \binits{T.}},
\bauthor{\bsnm{English}, \binits{Z.}},
\bauthor{\bsnm{Rombach}, \binits{R.}}:
\bctitle{Scaling rectified flow transformers for high-resolution image synthesis}.
In: \bbtitle{Forty-first International Conference on Machine Learning, {ICML} 2024, Vienna, Austria, July 21-27, 2024}
(\byear{2024}).
\bcomment{Accessed: 2025-02-13}.
\burl{https://openreview.net/forum?id=FPnUhsQJ5B}
\end{bchapter}
\endbibitem

\bibitem{Adobe}
\begin{botherref}
\oauthor{\bsnm{Adobe}}:
Adobe introduces Firefly Image 3 Foundation model to take Creative Exploration and ideation to new heights.
Accessed: 2025-02-13
(2024).
\url{https://news.adobe.com/news/news-details/2024/adobe-introduces-firefly-image-3-foundation-model-to-take-creative-exploration-and-ideation-to-new-heights}
\end{botherref}
\endbibitem

\bibitem{omnigen_t2i}
\begin{botherref}
\oauthor{\bsnm{Xiao}, \binits{S.}},
\oauthor{\bsnm{Wang}, \binits{Y.}},
\oauthor{\bsnm{Zhou}, \binits{J.}},
\oauthor{\bsnm{Yuan}, \binits{H.}},
\oauthor{\bsnm{Xing}, \binits{X.}},
\oauthor{\bsnm{Yan}, \binits{R.}},
\oauthor{\bsnm{Wang}, \binits{S.}},
\oauthor{\bsnm{Huang}, \binits{T.}},
\oauthor{\bsnm{Liu}, \binits{Z.}}:
Omnigen: Unified image generation.
CoRR
\textbf{abs/2409.11340}
(2024).
\doiurl{10.48550/ARXIV.2409.11340}
\end{botherref}
\endbibitem

\bibitem{stable_diffusion_3_5_large_t2i}
\begin{botherref}
\oauthor{\bsnm{Esser}, \binits{P.}},
\oauthor{\bsnm{Kulal}, \binits{S.}},
\oauthor{\bsnm{Blattmann}, \binits{A.}},
\oauthor{\bsnm{Entezari}, \binits{R.}},
\oauthor{\bsnm{M{\"{u}}ller}, \binits{J.}},
\oauthor{\bsnm{Saini}, \binits{H.}},
\oauthor{\bsnm{Levi}, \binits{Y.}},
\oauthor{\bsnm{Lorenz}, \binits{D.}},
\oauthor{\bsnm{Sauer}, \binits{A.}},
\oauthor{\bsnm{Boesel}, \binits{F.}},
\oauthor{\bsnm{Podell}, \binits{D.}},
\oauthor{\bsnm{Dockhorn}, \binits{T.}},
\oauthor{\bsnm{English}, \binits{Z.}},
\oauthor{\bsnm{Lacey}, \binits{K.}},
\oauthor{\bsnm{Goodwin}, \binits{A.}},
\oauthor{\bsnm{Marek}, \binits{Y.}},
\oauthor{\bsnm{Rombach}, \binits{R.}}:
Scaling rectified flow transformers for high-resolution image synthesis.
CoRR
\textbf{abs/2403.03206}
(2024).
\doiurl{10.48550/ARXIV.2403.03206}
\end{botherref}
\endbibitem

\bibitem{QK_normalization}
\begin{bchapter}
\bauthor{\bsnm{Henry}, \binits{A.}},
\bauthor{\bsnm{Dachapally}, \binits{P.R.}},
\bauthor{\bsnm{Pawar}, \binits{S.S.}},
\bauthor{\bsnm{Chen}, \binits{Y.}}:
\bctitle{Query-key normalization for transformers}.
In: \beditor{\bsnm{Cohn}, \binits{T.}},
\beditor{\bsnm{He}, \binits{Y.}},
\beditor{\bsnm{Liu}, \binits{Y.}} (eds.)
\bbtitle{Findings of the Association for Computational Linguistics: {EMNLP} 2020, Online Event, 16-20 November 2020}.
\bsertitle{Findings of {ACL}},
vol. \bseriesno{{EMNLP} 2020},
pp. \bfpage{4246}--\blpage{4253}.
\bpublisher{Association for Computational Linguistics},
\blocation{\hspace{0pt}}
(\byear{2020}).
\doiurl{10.18653/V1/2020.FINDINGS-EMNLP.379}
\end{bchapter}
\endbibitem

\bibitem{flux_schnell}
\begin{botherref}
\oauthor{\bsnm{Labs}, \binits{B.F.}}:
Announcing Black Forest Labs.
Accessed: 2025-01-11
(2024).
\url{https://blackforestlabs.ai/announcing-black-forest-labs/}
\end{botherref}
\endbibitem

\bibitem{diffusion_distillation}
\begin{bchapter}
\bauthor{\bsnm{Sauer}, \binits{A.}},
\bauthor{\bsnm{Lorenz}, \binits{D.}},
\bauthor{\bsnm{Blattmann}, \binits{A.}},
\bauthor{\bsnm{Rombach}, \binits{R.}}:
\bctitle{Adversarial diffusion distillation}.
In: \bbtitle{Computer Vision - {ECCV} 2024 - 18th European Conference, Milan, Italy, September 29-October 4, 2024, Proceedings, Part {LXXXVI}},
pp. \bfpage{87}--\blpage{103}
(\byear{2024}).
\bcomment{Accessed: 2025-02-13}.
\burl{https://doi.org/10.1007/978-3-031-73016-0\_6}
\end{bchapter}
\endbibitem

\bibitem{kolors}
\begin{botherref}
\oauthor{\bsnm{Team}, \binits{K.}}:
Kolors: Effective training of diffusion model for photorealistic text-to-image synthesis.
arXiv preprint
(2024)
\end{botherref}
\endbibitem

\bibitem{NeuralLoveAI_2024}
\begin{botherref}
\oauthor{\bsnm{AI}, \binits{N.L.}}:
Ai art revolution: Introducing auto-aesthetics for personalized gen AI experience.
Accessed: 2025-02-13
(2024).
\url{https://neural.love/blog/auto-aesthetics-v1-ai-art-revolution}
\end{botherref}
\endbibitem

\bibitem{CLIP}
\begin{bchapter}
\bauthor{\bsnm{Radford}, \binits{A.}},
\bauthor{\bsnm{Kim}, \binits{J.W.}},
\bauthor{\bsnm{Hallacy}, \binits{C.}},
\bauthor{\bsnm{Ramesh}, \binits{A.}},
\bauthor{\bsnm{Goh}, \binits{G.}},
\bauthor{\bsnm{Agarwal}, \binits{S.}},
\bauthor{\bsnm{Sastry}, \binits{G.}},
\bauthor{\bsnm{Askell}, \binits{A.}},
\bauthor{\bsnm{Mishkin}, \binits{P.}},
\bauthor{\bsnm{Clark}, \binits{J.}},
\bauthor{\bsnm{Krueger}, \binits{G.}},
\bauthor{\bsnm{Sutskever}, \binits{I.}}:
\bctitle{Learning transferable visual models from natural language supervision}.
In: \beditor{\bsnm{Meila}, \binits{M.}},
\beditor{\bsnm{Zhang}, \binits{T.}} (eds.)
\bbtitle{Proceedings of the 38th International Conference on Machine Learning, {ICML} 2021, 18-24 July 2021, Virtual Event}.
\bsertitle{Proceedings of Machine Learning Research},
vol. \bseriesno{139},
pp. \bfpage{8748}--\blpage{8763}.
\bpublisher{{PMLR}},
\blocation{\hspace{0pt}}
(\byear{2021}).
\bcomment{Accessed: 2025-02-13}.
\burl{http://proceedings.mlr.press/v139/radford21a.html}
\end{bchapter}
\endbibitem

\end{thebibliography}

\end{document}